\documentclass[letterpaper, 10 pt, journal, twoside]{ieeetran}

\IEEEoverridecommandlockouts                              

\usepackage{acro}
\usepackage{lineno}
\usepackage{graphicx}
\usepackage{amsmath}
\usepackage{amssymb}
\usepackage{booktabs}

\DeclareAcronym{ctslam}{
    short = CT-SLAM,
    long = Continuous Time Simultaneous Localization and Mapping
}
\DeclareAcronym{slam}{
    short = SLAM,
    long = Simultaneous Localization and Mapping
}
\DeclareAcronym{bp}{
    short = BP,
    long = Belief Propagation
}
\DeclareAcronym{gbp}{
    short = GBP,
    long = Gaussian Belief Propagation
}
\DeclareAcronym{gp}{
    short = GP,
    long = Gaussian Process,
    long-plural = es
}
\DeclareAcronym{ate}{
    short = ATE,
    long = Absolute Trajectory Error
}
\DeclareAcronym{are}{
    short = ARE,
    long = Absolute Rotation Error
}
\DeclareAcronym{sota}{
    short = SOTA,
    long = State-of-the-art
}
\DeclareAcronym{pgo}{
    short = PGO,
    long = Pose Graph Optimization
}
\DeclareAcronym{ipu}{
    short = IPU,
    long = Intelligent Processing Unit
}
\DeclareAcronym{cpu}{
    short = CPU,
    long = Central Processing Unit
}
\DeclareAcronym{gpu}{
    short = GPU,
    long = Graphics Processing Unit
}
\DeclareAcronym{ba}{
    short = BA,
    long = Bundle Adjustment
}
\DeclareAcronym{plgo}{
    short = PLGO,
    long = Pose-Landmark Graph Optimization
}
\DeclareAcronym{dl}{
    short = DL,
    long = Deep Learning
}
\DeclareAcronym{map}{
    short = MAP,
    long = Maximum A Posteriori
}
\DeclareAcronym{gn}{
    short = GN,
    long = Gauss-Newton
}
\DeclareAcronym{nlls}{
    short = NLLS,
    long = Non Linear Least Squares
}
\DeclareAcronym{ai}{
    short = AI,
    long = Artificial Intelligence
}
\DeclareAcronym{lidar}{
    short = LiDAR,
    long = Light Detection And Ranging
}
\DeclareAcronym{imu}{
    short = IMU,
    long = Inertial Measurement Unit
}
\DeclareAcronym{pba}{
    short = PBA,
    long = Photometric Bundle Adjustment
}

\usepackage{xcolor}
\usepackage{amsopn}
\usepackage{algpseudocode}
\usepackage{booktabs}
\usepackage{graphicx}
\usepackage{multirow}
\usepackage{siunitx}
\usepackage[table]{xcolor}
\usepackage{array} 
\usepackage{amsmath}
\usepackage{amssymb}
\usepackage{subcaption}
\usepackage{hyperref}

\definecolor{headerblue}{RGB}{220,230,241} 
\definecolor{rowgray}{RGB}{245,245,245} 

\def\figref#1{Fig.~\ref{#1}}

\def\eqref#1{Eq.~(\ref{#1})}

\newcommand{\Manifold}{\mathcal{M}}
\newcommand{\Tangent}[1]{\mathrm{T_{#1}}\Manifold}

\usepackage{algpseudocode}

\newcommand{\bR}{\mathbf{R}}

\newcommand{\bQ}{\mathbf{Q}}
\newcommand{\bT}{\mathbf{T}}

\newcommand{\bJ}{\mathbf{J}}

\newcommand{\cJ}{\mathcal{J}}
\newcommand{\cN}{\mathcal{N}}

\newcommand{\cP}{\mathcal{P}}

\newcommand{\bbR}{\mathbb{R}}
\newcommand{\bbSE}{\mathbb{SE}}

\newcommand{\be}{\mathbf{e}}

\newcommand{\bx}{\mathbf{x}}

\newcommand{\btau}{\boldsymbol{\tau}}
\newcommand{\bmu}{\boldsymbol{\mu}}
\newcommand{\bnu}{\boldsymbol{\nu}}
\newcommand{\bgamma}{\boldsymbol{\gamma}}
\newcommand{\bxi}{\boldsymbol{\xi}}
\newcommand{\bvarpi}{\boldsymbol{\varpi}}
\newcommand{\bomega}{\boldsymbol{\omega}}
\newcommand{\bfeta}{\boldsymbol{\eta}}
\newcommand{\bSigma}{\mathbf{\Sigma}}

\newcommand{\bPhi}{\mathbf{\Phi}}

\newcommand{\bLambda}{\mathbf{\Lambda}}

\DeclareMathOperator*{\argmin}{argmin}

\def\g2o{$g^2o$}
\def\t2v{\mathrm{log}}
\def\v2t{\mathrm{exp}}
\def\ev2t{\mathrm{ev2t}}

\def\figref#1{Fig.~\ref{#1}}

\def\eqref#1{Eq.~(\ref{#1})}

\author{Davide Ceriola$^{1,*}$, Simone Ferrari$^{1,*}$, Luca Di Giammarino$^{1}$, Leonardo Brizi$^{1,2}$, Giorgio Grisetti$^{1}$%
\thanks{$^{*}$These authors contributed equally to this work.}
\thanks{This work has been partially supported by PNRR MUR project PE0000013-FAIR.}
\thanks{$^{1}$D. Ceriola, S. Ferrari, L. Di Giammarino, L. Brizi, and G. Grisetti are with the Department of Computer, Control, and Management Engineering ``Antonio Ruberti", Sapienza University of Rome, Italy.
Email:\,\,{\tt\footnotesize{\{ceriola, s.ferrari, digiammarino, brizi, grisetti\}@diag.uniroma1.it.}}}
\thanks{$^{2}$L. Brizi is also with the University of Stuttgart, Germany.}
}

\title{Breaking Time: A Fully Gaussian Framework for Distributed and Continuous-Time SLAM}

\begin{document}

\maketitle
\begin{abstract}
Continuous-time SLAM provides a principled framework for fusing heterogeneous sensors while estimating smooth trajectories, and is particularly well-suited for handling heterogeneous, asynchronous sensor streams with non-uniform readout patterns, such as rolling shutter cameras, LiDAR scanners, radar sweeps, or event-based sensors.
In this work, we introduce G-solver, a fully Gaussian and distributed framework that combines Gaussian Belief Propagation (GBP) with Gaussian Process (GP) motion priors for continuous-time trajectory estimation. Our GP model provides a probabilistic representation of the trajectory, enabling consistent interpolation and the use of data-driven hyperparameters, while GBP offers a scalable message-passing formulation well-suited for decentralized settings. The resulting solver naturally extends to multi-camera scenarios without specialized synchronization or engineering effort. We evaluate the approach on synthetic and real data, including rolling shutter and distributed multi-camera optimization, demonstrating accurate and stable estimation with runtimes comparable to existing continuous-time methods. An open-source implementation is released at \mbox{\url{https://github.com/rvp-group/gsolver}}.
\end{abstract}

\section{Introduction}
\label{sec:introduction}
\IEEEPARstart{A}{ccurate} ego-motion estimation is central to many vision-based applications, from mobile AR to autonomous navigation. Modern systems fuse heterogeneous sensor modalities: cameras, IMUs, GNSS, LiDAR, and event cameras, to remain robust under motion blur, low texture, or extreme dynamics.
Most existing \ac{slam} pipelines \cite{campos2021orb, murai2025mast3r, lipson2024deep} adopt a discrete-time formulation. While effective, discrete-time models require explicit handling of temporal offsets and high-rate asynchronous sensors. \ac{ctslam} addresses these issues by representing trajectories as smooth functions of time, enabling natural fusion of heterogeneous measurements \cite{Barfoot2014BatchCT}.

\begin{figure}[t]
    \centering
\includegraphics[width=0.45\textwidth]{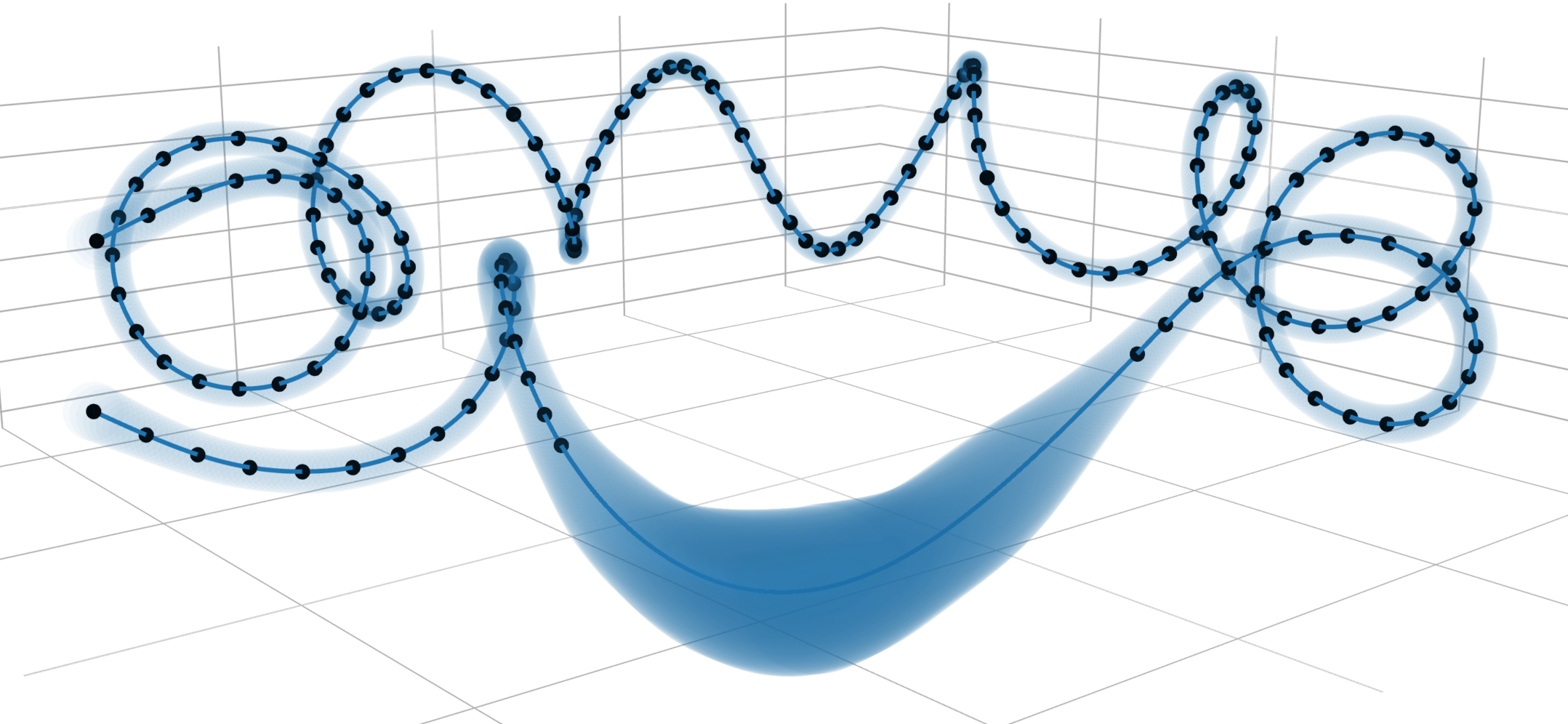}
\caption{\textbf{Querying continuous-time mean and covariance.} 
G-solver GP-based trajectory estimate on a \textit{torus} motion sequence. 
Shaded regions depict the posterior covariance, which naturally expands in segments 
with fewer observations and contracts where measurements are available. 
The model supports querying both trajectory mean and uncertainty at arbitrary timestamps.}
\label{fig:covariance}
\end{figure}

Trajectory representations in \ac{ctslam} often rely on spline models, where motion is parameterized by knots and associated basis functions. Splines provide smooth trajectories and efficient evaluation, but their behavior is governed by design choices, such as knot spacing, order, and basis structure, that must be fixed a priori. These choices implicitly determine the temporal resolution and motion assumptions of the model, and adjusting them to different motions or sensing conditions typically requires manual tuning. A second limitation is uncertainty handling: although spline uncertainty can be computed, propagating it through the full inference pipeline is less straightforward, making principled probabilistic fusion more challenging. \acp{gp} address these limitations by treating trajectories as distributions over functions rather than fixed basis expansions. \acp{gp} priors directly encode motion uncertainty, allow querying the trajectory at arbitrary timestamps, and expose hyperparameters that can often be tuned or learned from data. These properties make \acp{gp} a flexible and probabilistically consistent alternative for continuous-time estimation and multi-sensor fusion, particularly in scenarios with asynchronous measurements or varying sensor readouts.

\ac{ctslam} is typically solved using centralized \ac{nlls} optimization, either in batch or fixed-lag form \cite{talbot2025continuous}. Although successful, repeated relinearization and centralized computation become limiting when processing high-frequency asynchronous sensors, multi-camera systems, or rolling shutter cameras with row-wise timestamps.
SLAM problems can be represented as factor graphs, where variables are linked by probabilistic constraints. \ac{gbp} performs distributed inference on such graphs via local Gaussian message passing, providing an alternative to centralized optimization. \ac{gbp} supports incremental updates without requiring a global solve, making it well-suited for continuous-time estimation with heterogeneous or asynchronous sensor streams.

Building on continuous-time estimation \cite{Barfoot2014BatchCT} and distributed Gaussian message passing \cite{ortiz2023gaussian}, we introduce G-solver, a fully Gaussian framework that combines \acp{gp} and \ac{gbp} for distributed continuous-time SLAM. The approach handles asynchronous sensing, nonuniform readout, and multi-camera setups without additional synchronization mechanisms, providing a clean and general probabilistic formulation.
\section{Related Work}
\label{sec:related_work}
State estimation is a fundamental problem in robotics and computer vision, spanning \ac{slam} \cite{Brachmann2016}, mapping \cite{Triggs1999}, tracking \cite{lsdslam}, and navigation \cite{Jacob1999}. Classical pipelines typically formulate inference as \ac{map} estimation solved via \ac{nlls} optimization \cite{ceres2023, Kummerle11}. Discrete-time representations remain widely used, but they require explicit handling of temporal offsets, high-rate sensors, and multi-sensor synchronization, which motivates the consideration of continuous-time alternatives.

\paragraph{Continuous-Time Trajectory Models}
B-Spline trajectory representations \cite{Sommer2020} provide locality and smoothness, while Z-Spline \cite{zspline} provides a moment-conserving interpolation. However, both require temporal resolution and basis structure to be fixed a priori; while uncertainty can be recovered, propagation within a probabilistic pipeline is less direct.
Gaussian-process trajectory models \cite{Barfoot2014BatchCT, Anderson2015} embed uncertainty intrinsically and support data-driven hyperparameter learning. Prior \ac{gp} motion models, including constant-velocity, Singer \cite{wong2020data}, and ESGVI \cite{wong2020variational} formulations, demonstrate the flexibility of continuous-time priors. Recent comparisons \cite{johnson2024continuoustimetrajectoryestimationcomparative} show that GP-based inference is competitive in both accuracy and efficiency.

\paragraph{Optimization and Scalability}
Continuous-time SLAM is often solved through centralized \ac{nlls}, either in batch or fixed-lag form \cite{talbot2025continuous}. Although effective, centralized solvers struggle to scale when faced with asynchronous high-rate sensing, nonuniform readout mechanisms such as rolling shutters or scanning LiDARs, and multi-camera systems operating at mismatched frame rates. Large-scale problems \ac{ba} \cite{duisterhof2025mast3r} further expose memory and runtime constraints.

\paragraph{Distributed Inference and \ac{gbp}}
Distributed inference has gained attention for multi-robot SLAM \cite{Polizzi2022} and with the emergence of graph-parallel hardware such as \acp{ipu}. \ac{gbp} performs local Gaussian message passing on factor graphs \cite{Bishop2006}, offering exact inference on trees and strong empirical performance on loopy graphs \cite{Bickson2008}. Recent work \cite{JianConvAnalys, Davison2018, Davison2019, ortiz2023gaussian} highlights its scalability. Hyperion \cite{hyperion} applies \ac{gbp} to continuous-time SLAM via B- and Z-Splines. Our approach instead employs \ac{gp} priors within \ac{gbp}, providing uncertainty-aware continuous trajectories while enabling distributed inference. This is particularly advantageous for asynchronous multi-camera setups and rolling shutter data where row-level timestamps must be handled consistently.

\paragraph{Contributions}
Our main contributions are:
(i) A fully Gaussian continuous-time SLAM framework that integrates \ac{gp} motion priors within a \ac{gbp} inference scheme, enabling distributed optimization and naturally accommodating asynchronous multi-camera inputs and nonuniform sensor readout observations. (ii) A probabilistic trajectory representation in which the same \ac{gp} hyperparameters govern both the optimization of discrete states and the interpolation of continuous-time queries, ensuring internal consistency. (iii) An experimental evaluation on synthetic and real data showing accurate and stable estimation, effective uncertainty handling, and favorable scalability properties, together with an open-source implementation.

\section{G-solver}
\label{sec:ct-slam}
\begin{figure}[t]
    \centering
    \includegraphics[width=\columnwidth]{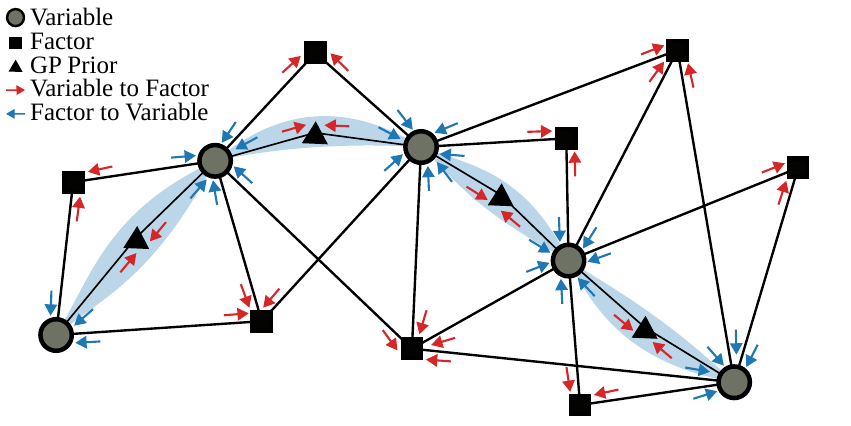}
    \caption{\textbf{G-solver: Continuous-time inference via message passing.} 
    A factor graph view of our method: Gaussian Belief Propagation messages (arrows) propagate local information across the graph, while Gaussian Process priors model the trajectory continuously in time. The light-blue region illustrates the GP uncertainty envelope.}
    \label{fig:motiv}
\end{figure}

We propose a \ac{gbp} solver built around \ac{gp} priors for continuous-time trajectory estimation in \ac{slam}. 
G-solver estimates the most likely trajectory by combining sensor measurements with 
continuous-time motion constraints (\figref{fig:motiv}). Combining \ac{gp} priors and \ac{gbp} requires several non-trivial design
choices for numerical stability, manifold consistency, and reliable
message-passing convergence.

Instead of relying on spline interpolation as in \cite{hyperion}, we adopt a constant-velocity 
\ac{gp} prior between consecutive poses. This prior is implemented as a binary \ac{gp} factor that 
constrains both pose and body-centric velocity, enabling smooth and uncertainty-aware continuous-time 
trajectory estimation. Two practical advantages follow from this formulation. First, the \ac{gp} prior exposes a 
small set of physically interpretable hyperparameters (the entries of $\bQ_C$), which can be tuned or learned 
from data, whereas spline-based approaches depend on fixed design choices such as knot spacing and spline order. 
Second, the \ac{gp} prior yields an explicit quadratic cost, allowing it to integrate cleanly into the 
\ac{gbp} machinery and directly influence the estimated trajectory through iterative message passing.

An additional and important advantage of our \ac{gp} formulation lies in the resulting factor graph structure. 
Spline-based continuous-time models, such as those used in Hyperion~\cite{hyperion}, couple each pose to multiple 
control points (e.g., four for a cubic B-Spline), which causes even simple measurement factors to introduce 
dense loops in the graph. These cyclic dependencies are known to challenge the convergence and stability of 
iterative \ac{gbp} \cite{JianConvAnalys}. In contrast, our \ac{gp} prior links only successive poses, producing a chain-like motion model 
that dramatically reduces the number of loops. Our experiments demonstrate that this topology is inherently more compatible with distributed 
message passing and leads to more predictable convergence behavior, especially in large-scale or 
multi-camera settings.

\paragraph{Preliminaries}
In our work, the estimated quantities include poses in $\bbSE(3)$, hence the formulation needs to support manifold-valued states. 
We denote the manifold as $\Manifold$ and the associated tangent space at $\bmu \in \Manifold$ as $\Tangent{\bmu}$. 
We rely on the $\boxplus$-notation as in \cite{hertzberg2013integrating}, defining the operators
$\boxplus : \Manifold \times \Tangent{\bmu} \mapsto \Manifold$ and 
$\boxminus : \Manifold \times \Manifold \mapsto \Tangent{\bmu}$,
which allows to move between the manifold and its tangent space. 
The rest of the notation follows \cite{Bishop2006}. In the remainder of this section, the inference problem is formulated as a factor graph.
Each variable node $x_i$ is associated with a Gaussian belief
\begin{equation}
    x_i \sim \cN(\bmu_{x_i}, \bSigma_{x_i}) 
    = \cN^{-1}(\bfeta_{x_i}, \bLambda_{x_i}),
\end{equation}
and each factor $f_j$ with an information-form Gaussian
\begin{equation}
    f_j \sim \cN^{-1}(\bfeta_{f_j}, \bLambda_{f_j}).
\end{equation}
The mean $\bmu \in \Manifold$ lies on the manifold, while the covariance 
$\bSigma = \bLambda^{-1} \in \bbR^{\mathrm{dim}(\Tangent{\bmu}) \times \mathrm{dim}(\Tangent{\bmu})}$ 
is defined in the corresponding tangent space. 
The information vector $\bfeta$ and precision matrix $\bLambda$ satisfy $\bfeta = \bLambda \bmu$.
\begin{figure}[t]
    \centering
    \includegraphics[width=0.95\columnwidth]{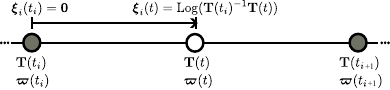}
    \caption{\textbf{Locally linear approximation} between each pair of measurement times, \( t_i \) and \( t_{i+1} \), where a constant velocity model is applied. This figure illustrates the relationships between the local pose variable \( \bxi_i(t) \), and the global trajectory state \( \{\bT(t), \bvarpi(t)\} \).}
    \label{fig:GP_prior}
\end{figure}

\subsection{Continuous time estimation using GP}
For continuous-time estimation, we introduce GP-based prior factors with their own quadratic cost. The cost depends on both pose and velocity, so the latent state must include both components. We model the trajectory as $\bx(t) \sim \mathcal{GP}(\bmu(t), \pmb{\mathcal{K}}(t, t^\prime))$, where $\bmu(t)$ and $\pmb{\mathcal{K}}(t, t^\prime)$ are the mean and covariance of the GP, respectively. The Markov state is defined as:
\begin{equation}
\bx(t) \doteq \{\bT(t), \bvarpi(t)\},
\label{eq:comp}
\end{equation}
where $\bT(t) \in \mathbb{SE}(3)$ represents the pose and $\bvarpi(t) \in \mathbb{R}^6$ is the body-centric velocity, consisting of both linear $\bnu(t)$ and angular $\bomega(t)$ components.

We define binary factors between consecutive poses to incorporate this \ac{gp} into the factor graph, encoding the constant velocity prior as a \ac{gp} factor, as shown in \figref{fig:GP_prior}. In contrast to \cite{Anderson2015}, which linearizes the pose increment entirely in the tangent space using right Jacobians, we evaluate the pose component of the error directly on the $\bbSE(3)$ manifold. This yields a slightly different discretization of the GP prior, chosen for improved numerical stability within our \ac{gbp} implementation. The corresponding error term for the binary \ac{gp} factor between poses $\bT_i$ and $\bT_{i+1}$ is:
\begin{equation}
\be_i = \begin{bmatrix} (\bT_{i}^{-1}\bT_{i+1}) \boxminus \mathrm{Exp}(\Delta t_i \bvarpi_i) \\
\begin{pmatrix} ^i\bR_{i+1} & \mathbf{0} \\
\mathbf{0} & ^i\bR_{i+1} \end{pmatrix} \bvarpi_{i+1} - \bvarpi_i \end{bmatrix},
\label{eq:gp_error}
\end{equation}
where $\bT_i, \bT_{i+1}, \bvarpi_i, \bvarpi_{i+1}$ are the global state variables, while $\Delta t_i = t_{i+1} - t_i$ and $^i\bR_{i+1}$ are respectively time and rotational difference between  consecutive poses. We found this formulation to be more numerically stable than a purely tangent-space linearization for larger inter-frame motions. The error term 
$\be_i$ penalizes deviations from a body-centric constant-velocity trajectory and serves as the \ac{gp} motion prior. The \textit{local} Markov state is defined by 
\begin{equation}
\bgamma_i(t) \doteq \begin{bmatrix} \bxi_i(t) \\ \dot{\bxi_i}(t) \end{bmatrix},
\end{equation}
where $\bxi_i(t) = \mathrm{Log}(\bT_i^{-1} \bT(t))$ is the local pose variable, defined in the tangent space of the manifold $\mathbb{SE}(3)$, and $\dot{\bxi_i}(t)$ represents the velocity.  Differently from other approaches \cite{Anderson2015, 8461077}, we map the body-centric twist $\bvarpi(t)$ into the local frame of $\bT_i$ using only the relative rotation,
\begin{equation}
\dot{\bxi_i}(t) = \begin{pmatrix} ^i\bR_t & \mathbf{0} \\
\mathbf{0} & ^i\bR_t \end{pmatrix} \bvarpi(t), \qquad t_i \leq t \leq t_{i+1},
\end{equation}
where $^i\bR_t = \bR_i^\top \bR_t$. This replaces the right-Jacobian mapping with a simpler rotation-based
approximation, empirically stable for the considered motions. \\
As in \cite{Anderson2015}, the transition between states is governed by the constant-velocity model:
\begin{equation}
\bgamma_i(t_{i+1}) = \bPhi(t_{i+1}, t_i) \bgamma_i(t_i),
\end{equation}
where $\bPhi(t_{i+1}, t_i)$ is the transition matrix given by:
\begin{equation}
\bPhi(t_{i+1}, t_i) = \begin{bmatrix} \mathbf{I}_{6\times6} & (t_{i+1}-t_i) \mathbf{I}_{6\times6} \\
\mathbf{0} & \mathbf{I}_{6\times6} \end{bmatrix}.
\end{equation}
$\dot{\bvarpi}(t)$ is modeled as a zero-mean white-noise \ac{gp}:
\begin{equation}
\dot{\bvarpi}(t) \sim \mathcal{GP}(\mathbf{0}, \bQ_C\delta(t-t^\prime))
\end{equation}
where $\bQ_C$ is the power spectral density matrix of the process noise \cite{Barfoot2014BatchCT}. In practice, $\bQ_C$
(or a small set of scalar parameters defining it) can be tuned or learned from data \cite{Anderson2015, anderson2015batch}.
By integrating this process twice, we obtain the prior covariance matrix:
\begin{equation}
\bQ_i = \begin{bmatrix} \frac{1}{3} \Delta t_i^3 \bQ_C &
\frac{1}{2} \Delta t_i^2 \bQ_C \\ 
\frac{1}{2} \Delta t_i^2 \bQ_C &
\Delta t_i \bQ_C \end{bmatrix}, \qquad \Delta t_i = t_{i+1} - t_i.
\end{equation}
Finally, each \ac{gp} binary prior factor has cost:
\begin{equation}
E_i = \frac{1}{2} \be_i^\top \bQ_i^{-1} \be_i.
\label{eq:gp_energy}
\end{equation}

\subsection{Gaussian Belief Propagation with GP}
This section provides a detailed description of our solver, emphasizing its key differences from existing methods. \ac{gbp} is a decentralized iterative algorithm based on message passing \cite{JiangNIPS2011}. Each received message is internally remapped to the Euclidean tangent space to
operate on the underlying Gaussian distribution.  Let $m_{in} \sim \cN(\bmu_{in}, \bLambda_{in}^{-1})$ be a generic incoming message, we need to obtain the corresponding parameters on the tangent space $m_{in}^{\btau} \sim \cN(\btau_{in}, (\bLambda_{in}^{\btau})^{-1})$. Using the $\boxplus$ notation, this can be written as:
\begin{align}
    \btau_{in} &= \bmu_{in} \boxminus \bmu_{0}, \\
    \bLambda_{in}^{\btau} &\approx \bLambda_{in} \label{eq:lambdain}.
\end{align} 
We approximate the projected precision $\bLambda_{in}^{\btau} \approx \bLambda_{in}$, instead of explicitly computing the Jacobian-based first-order mapping \cite{hyperion, ortiz2023gaussian}. This well-established simplification provided a good efficiency/stability trade-off in all our experiments \cite{barfoot2024state, sola2018micro}. We recall that within the nodes, the internal computation is conducted on the Euclidean tangent space
as $\btau_{in} = (\bxi_i(t)^\top, \dot{\bxi_i}(t)^\top)^\top \in \mathbb{R}^{12}$. This mechanism naturally generalizes to Euclidean entities
such as point landmarks in $\mathbb{R}^3$, whose tangent space coincides with the domain. With this formulation, we can directly address a broader class of problems such as \ac{ba}.

\ac{gbp} consists of four main steps: the variable update, the factor update, the variable-to-factor message generation, and the factor-to-variable message generation. The first three steps are similar to the one proposed by \cite{ortiz2023gaussian}. We now discuss the factor-to-variable message since it differs from previous formulations \cite{hyperion}.

\paragraph{Factor-to-Variable Message}
\label{sec:f2v_message}
A factor-to-variable message encodes the probability distribution of the recipient variable that minimizes the factor cost function. To produce such a message, we first condition the factor to obtain the joint probability distribution over all the connected variables and then marginalize out the recipient variable. Hence, denoting by $\alpha$ the information related to the receiving variable and by $\beta$ the information about all other variables, the first step is encoded as:
\begin{align}
    \bfeta_{C_f} &= \begin{pmatrix} \bfeta_\alpha \\ \bfeta_\beta \end{pmatrix} 
        = \begin{pmatrix} \bfeta_{f_\alpha} \\ \bfeta_{f_\beta} + \bfeta_{x_\beta \rightarrow f_j} \end{pmatrix}, \\
    \bLambda_{C_f} &= \begin{bmatrix} \bLambda_{\alpha\alpha} & \bLambda_{\alpha\beta} \\ \bLambda_{\beta\alpha} & \bLambda_{\beta\beta} \end{bmatrix}
        = \begin{bmatrix} \bLambda_{f_{\alpha\alpha}} & \bLambda_{f_{\alpha\beta}} \\ \bLambda_{f_{\beta\alpha}} & \bLambda_{f_{\beta\beta}} + \bLambda_{x_\beta \rightarrow f_j} 
    \end{bmatrix},
\end{align}
where $\cN^{-1}(\bfeta_{f}, \bLambda_{f})$ is the factor distribution and $\cN^{-1}(\bfeta_{C_f}, \bLambda_{C_f})$ is the resulting conditioned joint distribution.
\begin{figure}[t]
    \centering
    \includegraphics[width=0.45\textwidth]{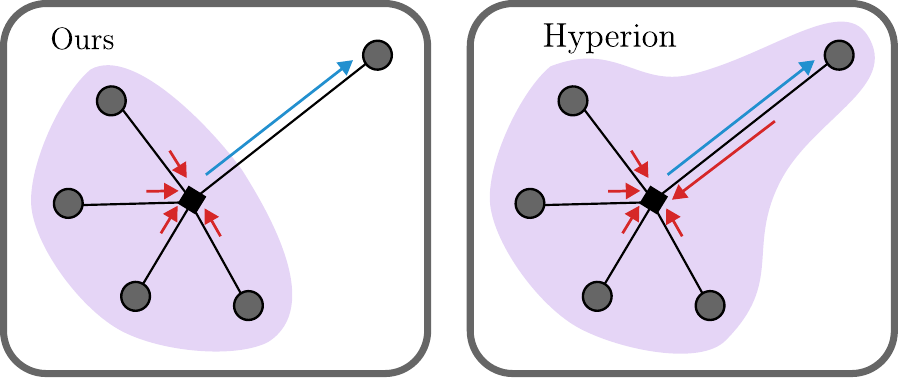}
    \caption{\textbf{Factor-to-Variable Message conditioning.} The illustration highlights the nodes (purple region) involved in computing the factor-to-variable message (light blue arrow). In Hyperion (right), the receiving variable is included in the conditioning set, whereas in our approach (left) it is excluded to avoid introducing bias into the computation.
    }
    \label{fig:conditioning}
\end{figure}
As shown in \figref{fig:conditioning}, unlike Hyperion \cite{hyperion}, our approach retains the original formulation by omitting the recipient variable from the conditioning set.

Following \cite{ortiz2023gaussian}, we compute the marginalization by using the Schur complement:
\begin{align}
    \bfeta_{M\alpha} &= \bfeta_\alpha - \Lambda_{\alpha \beta} \Lambda_{\beta \beta}^{-1} \bfeta_\beta, \\
    \Lambda_{M\alpha} &= \Lambda_{\alpha \alpha} - \Lambda_{\alpha \beta} \Lambda_{\beta \beta}^{-1} \Lambda_{\beta \alpha}.
\end{align}
The outward message $m_{f_j \rightarrow x_r} \sim \cN (\bmu_{f_j \rightarrow x_r}, \bLambda_{f_j \rightarrow x_r}^{-1})$ to the receiving variable $x_r$ can be retrieved by projecting the marginal distribution $\cN^{-1} (\bfeta_{M\alpha}, \bLambda_{M\alpha})$ back to the manifold.

Whereas each factor minimizes only its own energy, the \ac{gbp} machinery focuses on minimizing the total cost function:
\begin{align}
    \mathrm{E}_\mathrm{tot} &= \sum_{j \in \cJ}\frac{1}{2} \lVert \be_{j} \rVert^2_{\bLambda_{j}} + \sum_{i \in \cP}\frac{1}{2} \lVert \be_{i} \rVert^2_{\bQ_i^{-1}}.
\label{eq:tot_energy}
\end{align}
Here, the first sum encodes the energy of a generic set $\cJ$ of different measurement factors (priors, odometry, projective camera measurement, etc.), and the second captures the \ac{gp} prior costs $\cP$ defined in \eqref{eq:gp_energy}.

\begin{figure}[t]
    \centering
    \includegraphics[width=0.5\textwidth]{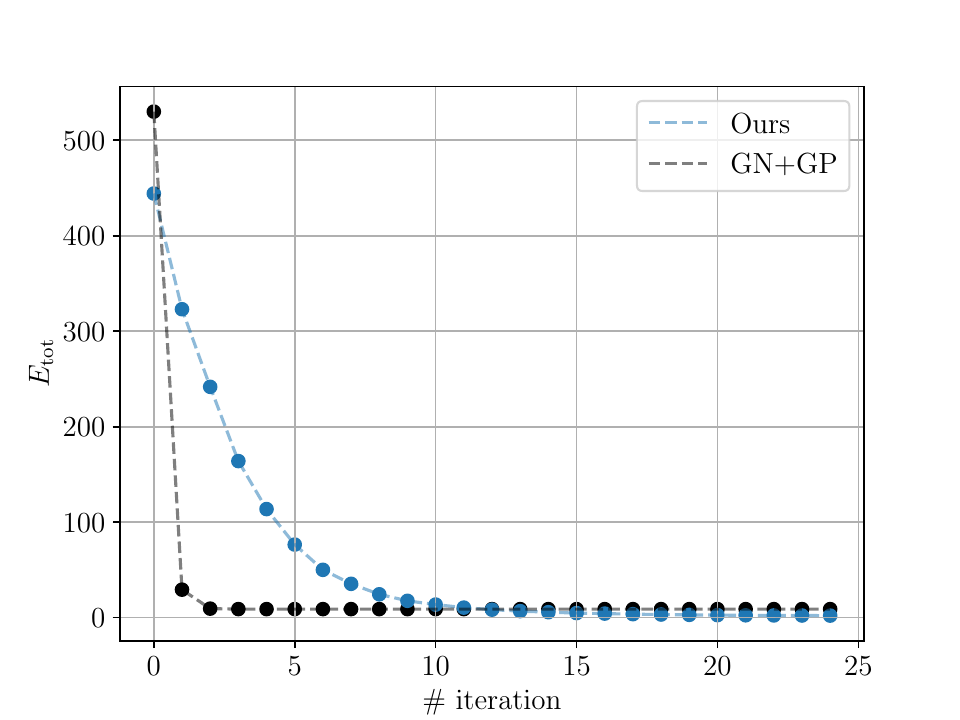}
    \caption{\textbf{Total energy over iterations.} G-solver and \ac{gn}+\ac{gp} total energy (\eqref{eq:tot_energy}) while solving PGO in the \textit{Helix} sequence, with initial perturbation and noise levels $\sigma=0.1$, $\eta_{ig}=1$ (both $\mathrm{[m]}$ and $\mathrm{[rad]}$). A G-solver iteration ends when every node has been updated and exchanged messages with all its neighbors.
    }
    \label{fig:gbp-gn}
\end{figure}

\begin{figure}[t]
    \centering
    \includegraphics[width=0.45\textwidth]{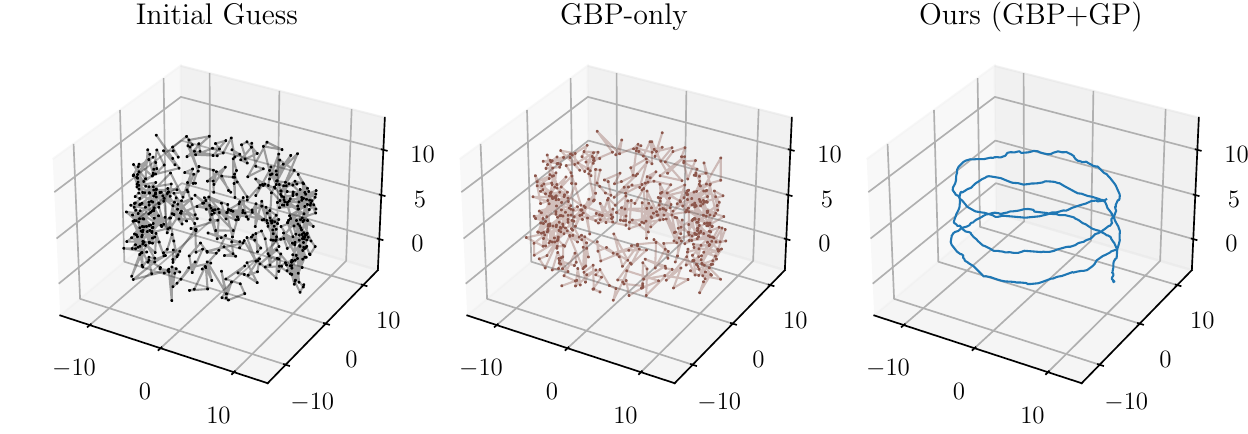}
    \caption{\textbf{Ablation on \ac{gp}.} Starting from a highly perturbed initial configuration (left), the trajectory optimized using only \ac{gbp} (middle) remains highly inaccurate, whereas (right) G-solver successfully recovers the \textit{helix} shape thanks to \ac{gp} priors. Prior-based example with $\sigma = 1$, $\eta_{ig} = 1$ (both $\mathrm{[m]}$ and $\mathrm{[rad]}$). }
    \label{fig:three_images}
\end{figure}
\section{Experiments}
\label{sec:experiments}
We evaluate G-solver on a range of problems requiring continuous-time trajectory estimation, including ChArUco-based \ac{ba}, rolling shutter cameras, and decentralized multi-camera setups. We validate three properties: (i) accurate continuous-time state estimation, (ii) computational efficiency comparable to state-of-the-art continuous-time optimizers, and (iii) the ability to handle asynchronous, distributed, and heterogeneous sensing without engineering overhead. A key advantage of our GP-based formulation is that the prior covariance is governed by a small and interpretable set of hyperparameters (encoded in $\bQ_{C}$), which can be inferred from data. This contrasts with spline-based approaches that rely on fixed basis functions, knot spacing, and spline order. Moreover, our use of \ac{gp} regression allows us to continuously query the posterior trajectory and its uncertainty after optimization (\figref{fig:covariance}), following the smoothing framework described in \cite{barfoot2024state}. Across the experiments, we evaluate accuracy using \ac{ate} and \ac{are} RMSE, and, where possible, reprojection error to assess robustness to asynchronous measurements. We also report computational metrics, such as runtimes and memory usage.
All evaluations were performed on a machine with an Intel(R) Core(TM) i9-14900KF processor (16 physical cores). We compare our method against the Hyperion solver~\cite{hyperion}, which represents trajectories using cubic Z-Splines and B-Splines and performs inference via distributed Gaussian message passing. In every experiment, both G-solver and Hyperion are configured to use a synchronous schedule for message exchanges between factors and variables.

\subsection{Synthetic Experiments}

\begin{figure}[t]
    \centering
    \includegraphics[width=0.45\textwidth]{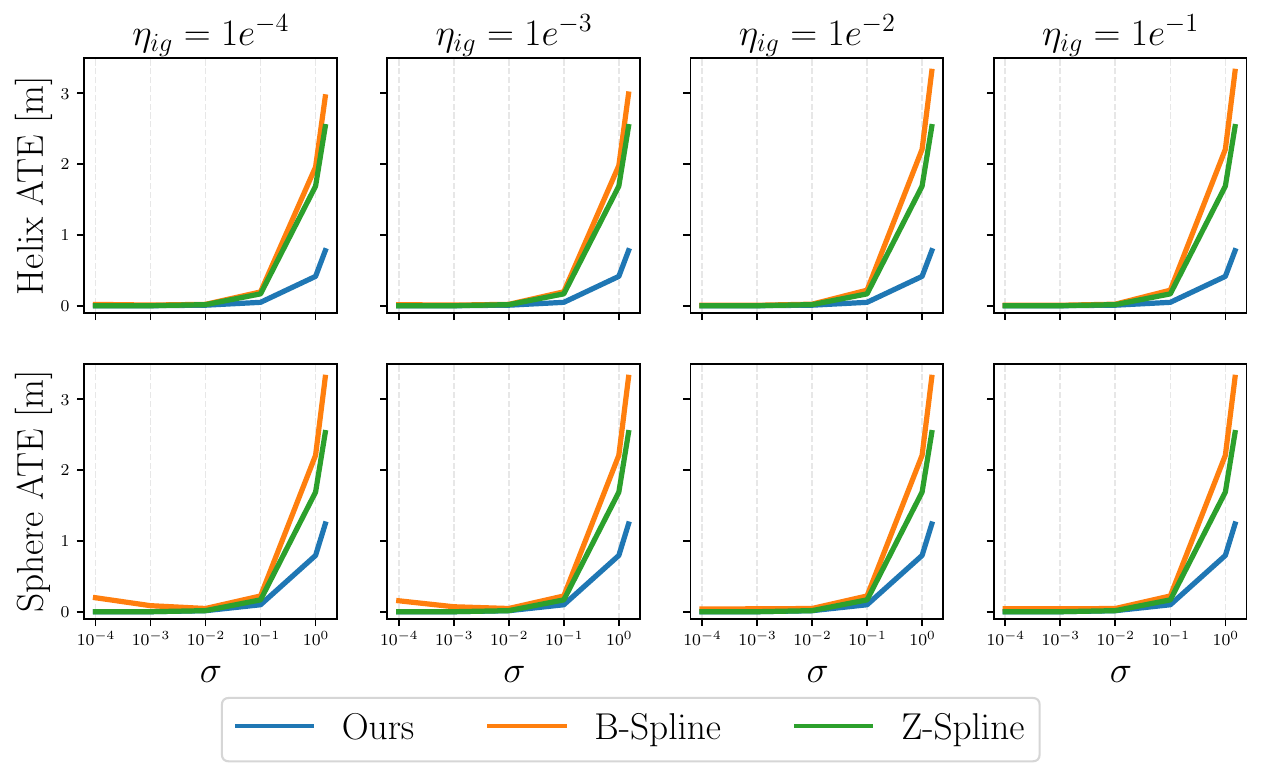}
    \caption{\textbf{Prior-based experiments.} Comparison of \textit{helix} and \textit{sphere} trajectories under varying noise levels with G-solver and Hyperion.}
    \label{fig:prior}
\end{figure}

\begin{figure}[t]
    \centering
    \includegraphics[width=0.45\textwidth]{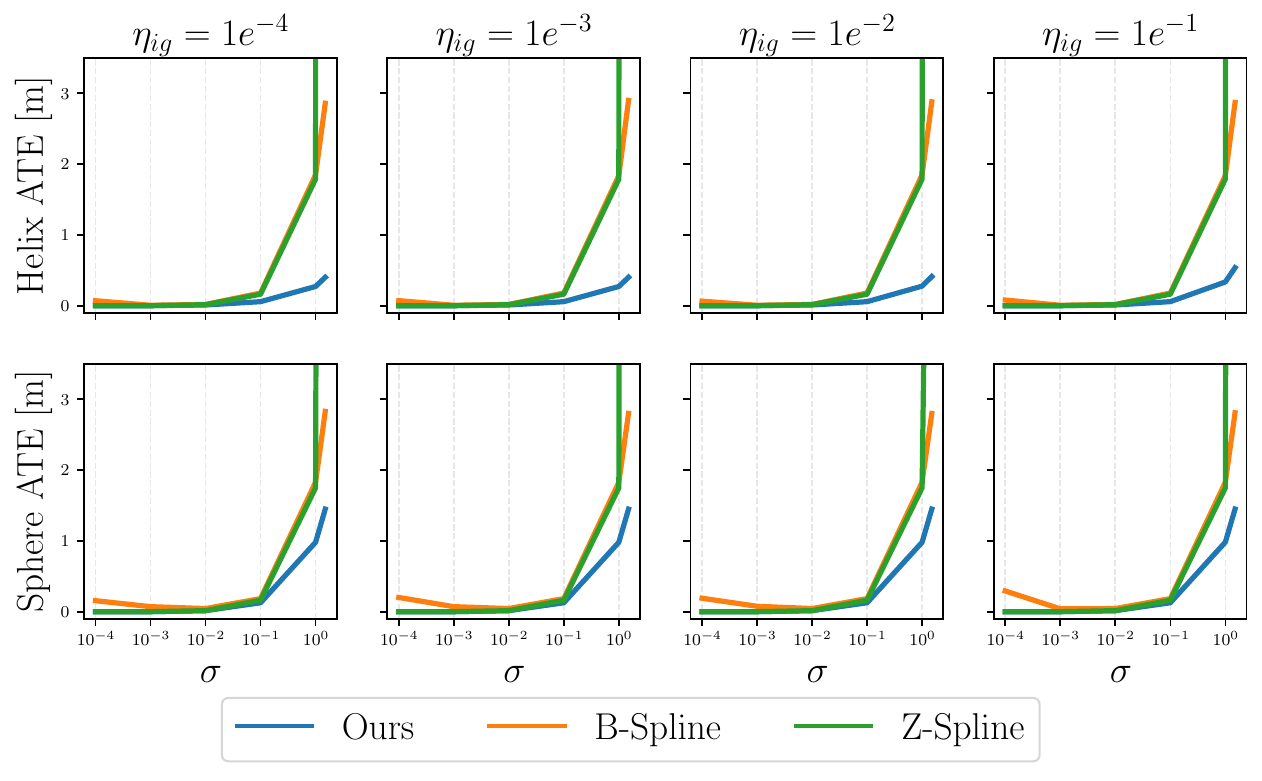}
    \caption{\textbf{\ac{pgo} experiments.} Comparison of \textit{helix} and \textit{sphere} trajectories under varying noise levels with G-solver and Hyperion.}
    \label{fig:pgo}
\end{figure}

We benchmark G-solver on two synthetic trajectories: \textit{helix} and \textit{sphere} that excite all six degrees of freedom and provide different translational and rotational motion profiles. Initial guess of the state estimate and measurements are perturbed respectively with anisotropic noise ($\mathrm{m}$ and $\mathrm{rad}$) $\eta_{ig}$ ranging from $10^{-4}$ to $10^{-1}$ and $\sigma$ ranging from $10^{-4}$ to $1.5$. We evaluate two tasks: prior-based localization using only absolute pose measurements in \figref{fig:prior}, and \ac{pgo} in \figref{fig:pgo}. To fairly compare continuous-time representations, all methods query the trajectory and covariance at a resolution $100\times$ finer than the control-point spacing. State initialization for G-solver uses finite-differencing of the initial pose sequence to generate twist estimates; Hyperion~\cite{hyperion} initializes their spline trajectories directly from the same noisy input. GP hyperparameters are initialized from a nominal reference value.  As shown in the subsequent ablation, G-solver remains robust even when these hyperparameters are perturbed by orders of magnitude. Finally, in \figref{fig:gbp-gn} we compare the energy over iterations of our solver and \ac{gn}.

\begin{figure} [t]
    \centering
    \includegraphics[width=0.45\textwidth]{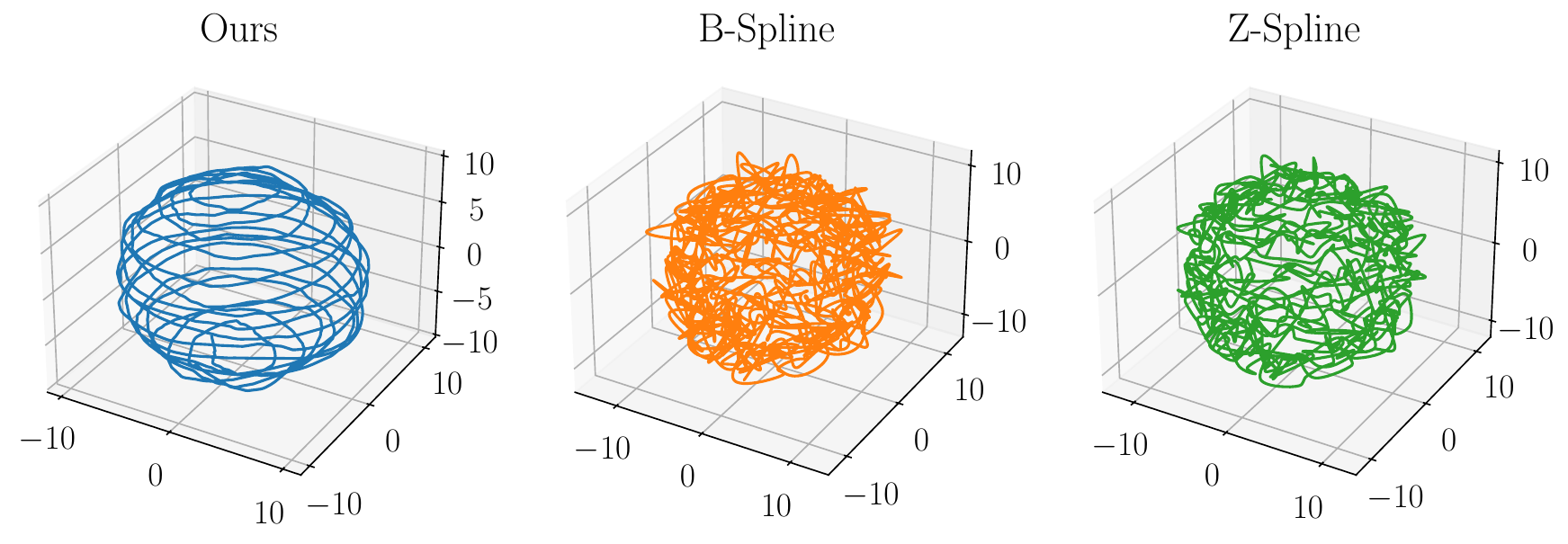}
    \caption{\textbf{Prior-based qualitative interpolation results.} Posterior continuous-time trajectory estimates for G-solver and Hyperion (B and Z spline) for the \textit{sphere} sequence, under significant perturbation and noise levels with $\sigma=1$, $\eta_{ig}=1$ (both $\mathrm{[m]}$ and $\mathrm{[rad]}$).}
    \label{fig:ours-hyperion}
\end{figure}

\begin{figure}[t]
    \centering
    \includegraphics[width=0.48\textwidth]{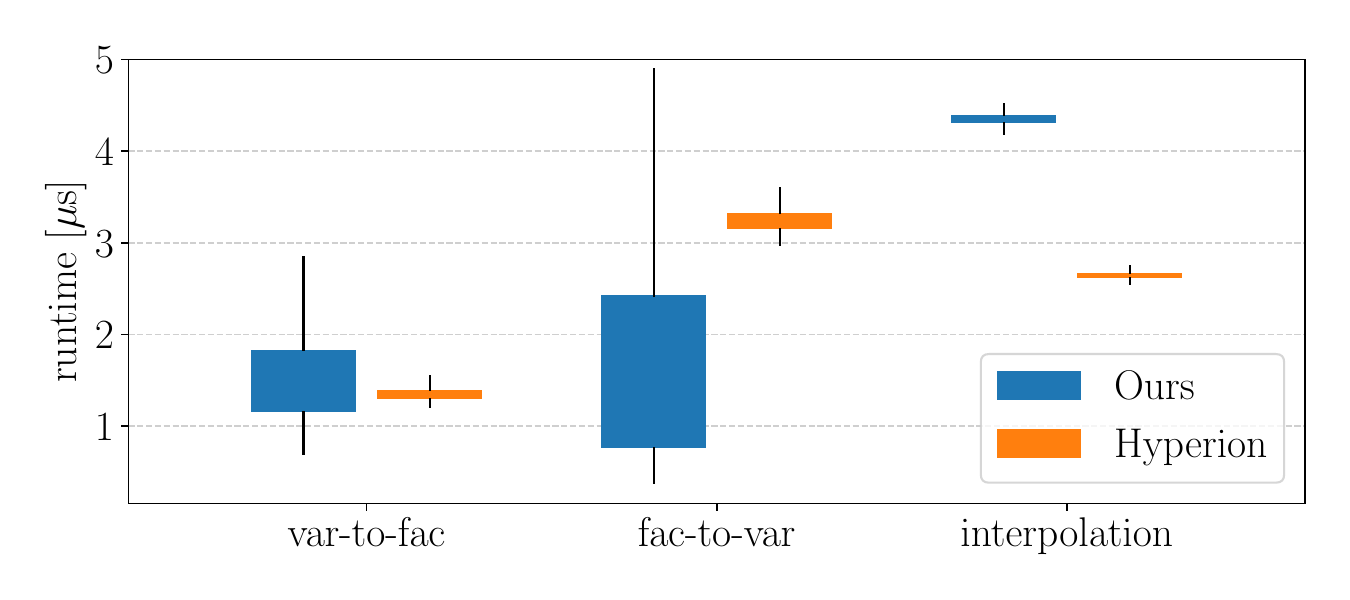}
    \caption{\textbf{Runtimes.} Runtime comparison between G-solver and Hyperion for message updates and continuous-time interpolation.}
    \label{fig:runtimes}
\end{figure}

\paragraph{Prior-Based Localization and Pose Graph Optimization}
Results for the \textit{helix} and \textit{sphere} trajectories (\figref{fig:prior} and \figref{fig:pgo}) reveal a consistent pattern. For low noise levels, all approaches achieve comparable \ac{ate} and \ac{are} RMSE values. In the lowest-noise regime, Z-Splines are slightly more accurate than G-solver, as their spline representations do not enforce a constant-velocity assumption. B-Splines, however, perform worse due to their approximating nature, which does not guarantee that the trajectory passes through the control points. As a result, the \ac{gp} prior in~\eqref{eq:gp_error} introduces a small residual cost even when measurements are nearly perfect. As noise increases, however, G-solver yields significantly lower error, often by an order of magnitude. In this regime, the \ac{gp} prior provides beneficial local smoothing that mitigates the impact of high measurement noise. \figref{fig:ours-hyperion} illustrates this behavior: G-solver remains stable and accurate under strong perturbations, while spline-based estimations become less robust. We also report runtime comparisons in \figref{fig:runtimes}. The results show that \ac{gp}-based continuous-time representation enables efficient interpolation. Combined with the locality of \ac{gbp} message passing, this yields an
online-capable solver with efficiency competitive with Hyperion.

\paragraph{G-Solver vs.\ Gauss--Newton with GP Priors}
\figref{fig:gbp-gn} compares G-solver with a \ac{gn} baseline equipped with the same \ac{gp} prior. Both methods converge to similar energy levels, but \ac{gn} converges faster because it performs a centralized batch update that solves the full linear system at each iteration. 
In contrast, G-solver relies on distributed message passing, where information propagates locally through the 
factor graph, resulting in slower global agreement, even though each factor update internally uses 
second-order information. On loopy graphs, such as those in \ac{pgo}, redundant cycles can induce oscillations, slowing convergence. This effect is reduced in prior-only problems, where the factor graph is closer to a chain and message propagation is more stable. Still, \ac{gbp} suits large, distributed, or multi-agent systems where centralized \ac{gn} updates are impractical. We further show in \figref{fig:three_images} that the inclusion of the \ac{gp} prior is critical under strong noise: without it, \ac{gbp} alone cannot recover the underlying trajectory, while G-solver reconstructs the shape robustly.

\paragraph{Consistency Analysis via NEES}
We assess the statistical consistency of the estimator through the Normalized
Estimation Error Squared (NEES), computed on the same Prior-Based Localization
problem and averaged over the \textit{helix} and \textit{sphere} trajectories
at increasing noise $\sigma$ (Tab.~\ref{tab:nees_evaluation}). For our 6-DOF
state, $6.0$ denotes an ideally calibrated covariance, with lower and higher
values indicating pessimistic and overconfident estimates. At realistic noise
($\sigma \ge 10^{-2}$), the mean NEES stays close to ideal and slightly
pessimistic, a desirable trait for robust estimation, with only the
\textit{sphere} at $\sigma = 1$ marginally exceeding it. As noise vanishes
($\sigma = 10^{-4}$), the estimation error nearly disappears while the GP prior
retains a baseline uncertainty, exhibiting a low NEES and indicating an overly
conservative estimate.

\begin{table}[h]
\centering
\scriptsize
\setlength{\tabcolsep}{4pt}
\begin{tabular}{@{}cccccc@{}}
\toprule
\multirow{2}{*}{Trajectory} & \multicolumn{5}{c}{$\sigma$} \\
\cmidrule(lr){2-6}
 & {1e-04} & {1e-03} & {1e-02} & {1e-01} & {1e-00} \\
\midrule
Helix  & 0.030 & 1.533 & 3.925 & 4.510 & 4.694 \\
Sphere & 0.030 & 1.542 & 4.187 & 4.589 & 6.515 \\
\bottomrule
\end{tabular}
\caption{\textbf{NEES evaluation.} Mean NEES for the Helix and Sphere trajectories across increasing noise levels $\sigma$.}
\label{tab:nees_evaluation}
\end{table}

\begin{figure}[t]
    \centering
\includegraphics[width=0.5\textwidth]{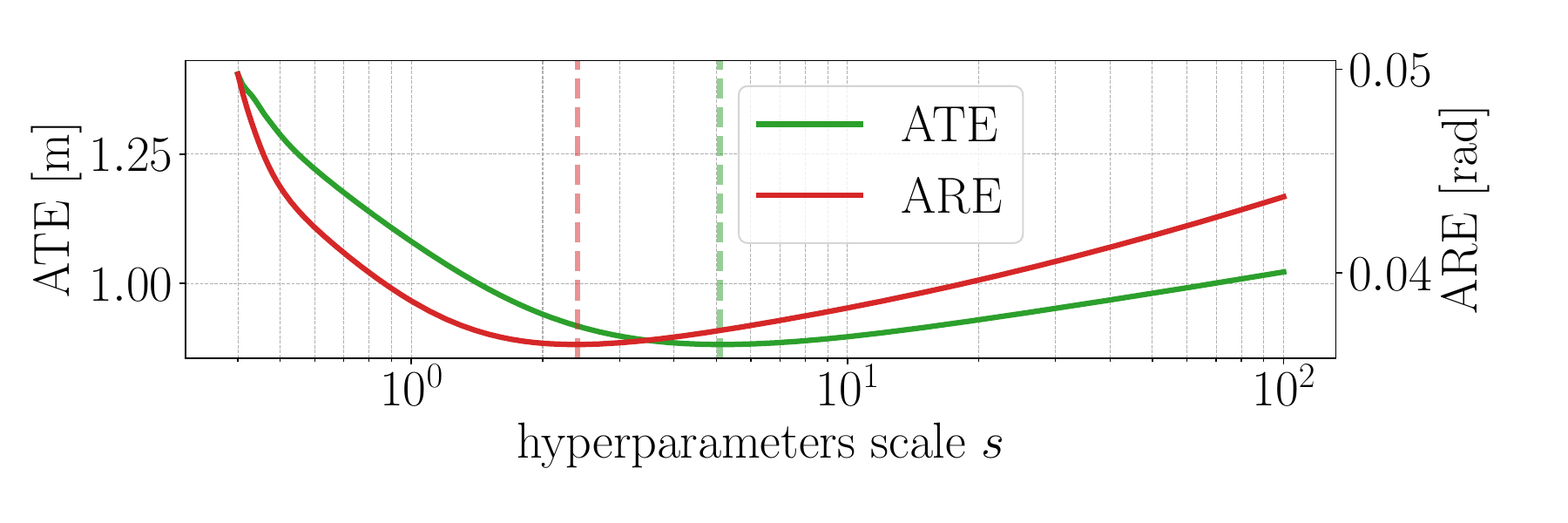}
\caption{\textbf{Ablation on hyperparameters.} ATE and ARE as a function of a scale factor $s$ applied to the learned $\bQ_C^\ast$. The metrics are minimized between $s=1$ and $s=10$, but performance remains stable over a wide range.}
    \label{fig:hyperparams}
\end{figure}

\paragraph{Ablation on Hyperparameters}
The \ac{gp} prior models body-centric accelerations as zero-mean white noise with covariance $\bQ_C$, whose diagonal entries represent acceleration variances~\cite{Anderson2015, anderson2015batch}.
A single $\bQ_C$ does not generalize across all motion profiles (e.g., wheeled versus legged locomotion), but tends to transfer effectively among platforms with similar dynamics. To assess sensitivity, we compute optimal hyperparameters $\bQ_C^\ast$ for a \textit{torus} trajectory and scale them by a factor $s$, i.e., $\bQ_C = s\bQ_C^\ast$. \figref{fig:hyperparams} shows \ac{ate} and \ac{are} as scale $s$ varies over several orders of magnitude. Performance peaks between $s = 1$ and $s = 10$. As expected, the optimal errors do not occur exactly at $s = 1$, since the accelerations along the \textit{torus} trajectory are not constant. Nevertheless, performance degrades only moderately when scale $s$ varies by an order of magnitude, demonstrating robustness to hyperparameter perturbations and indicating that an information-conservative approach to manual tuning can still make the method practical.
\begin{figure}[t]
    \centering
    \includegraphics[width=0.9\linewidth]{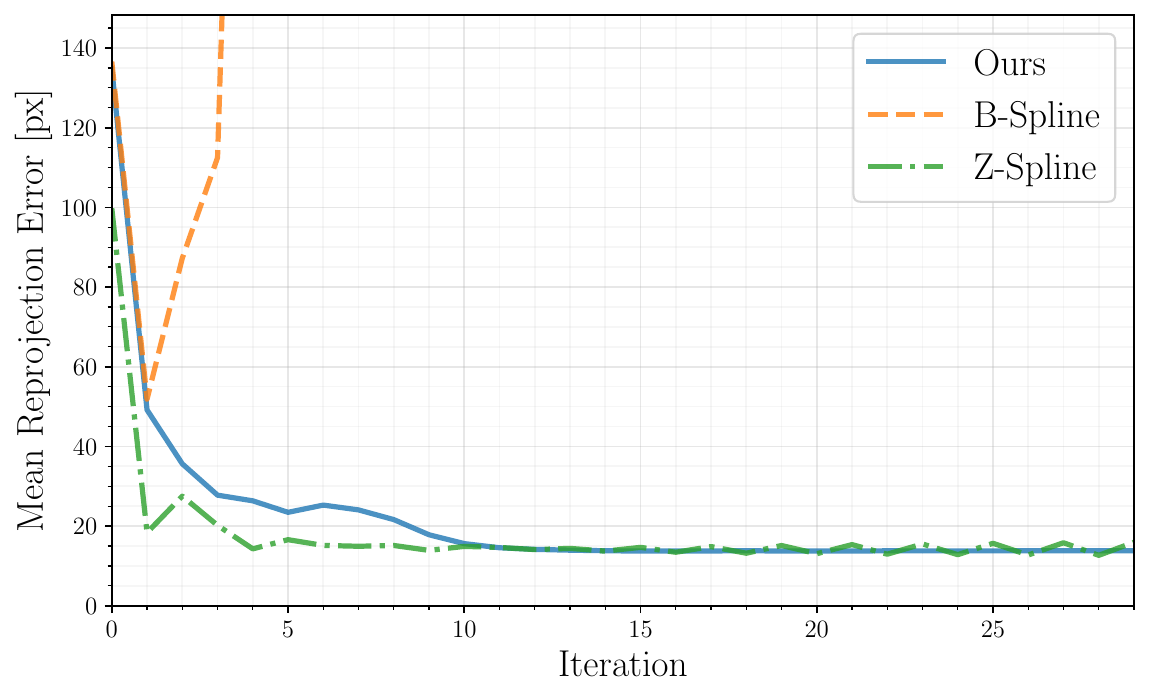}
    \caption{\textbf{ChArUco reprojection error.} Comparison of G-solver and Hyperion for real-world \ac{ba} using the mean of the reprojection error.}
    \label{fig:mean_reproj_error}
\end{figure}

\subsection{Real-World Use Cases}
To evaluate the robustness and flexibility of our approach, we consider three real-world use cases: ChArUco \ac{ba}, rolling shutter \ac{ba}, and multi-camera \ac{ba}. The Charuco sequence was captured with a smartphone camera moving in a room containing multiple ChArUco targets, with UcoSLAM \cite{munoz2020ucoslam} providing pose-landmark measurements as the front-end. For the rolling shutter and multi-camera scenarios, we use sequences from the KITTI dataset \cite{geiger2013vision}, with ORB-SLAM3 \cite{campos2021orb} serving as the front-end. These external systems are employed because our contribution focuses on the solver itself rather than a complete SLAM pipeline. In the experiments, B-Spline parametrization introduces instabilities that lead to diverging optimization. Such behavior is consistent with observations in \cite{hyperion}, where the destabilizing effect of B-Splines is attributed to the fact that they approximate the control points rather than interpolating them.

\paragraph{ChArUco BA}
In this setting, the factor graph includes pose variables as defined in \eqref{eq:comp}, while scene landmark variables are represented as $\mathbb{SE}(3)$ rigid bodies. Pose-landmark measurements from UcoSLAM \cite{munoz2020ucoslam} are used to jointly optimize the trajectory and map. The factor graph is initialized by adding Gaussian noise ($\sigma=0.1$) to the state estimates from the SLAM pipeline. The same camera calibration employed in UcoSLAM is used to compute the mean reprojection error over iterations, shown in \figref{fig:mean_reproj_error}. We find similar results to \cite{hyperion} as splines tend to be unstable in such visual scenarios. In fact, with B-Splines, which show a more volatile behavior, Hyperion diverges, while with Z-Splines, although it converges faster than G-solver, it exhibits oscillations after reaching the minimum reprojection error. G-solver, in contrast, converges smoothly and stabilizes after the 10th iteration.

\paragraph{Rolling shutter BA}
We evaluate our method on the KITTI benchmark \cite{geiger2013vision}, using sequence~\textit{06} as the validation sequence. To emulate rolling-shutter effects with controlled readout times, we reproject the ground truth trajectory to warp the stereo measurements produced by ORB-SLAM3 \cite{campos2021orb}. \ac{gp} hyperparameters for G-solver are initialized from a reference sequence (KITTI~\textit{05}). Both G-solver and Hyperion \cite{hyperion} optimize a set of per-image pose variables, where each pose corresponds to a block of image rows. 
All poses associated with a given image are initialized to the corresponding estimate provided by ORB-SLAM3, meaning that larger simulated readout times result in a noisier and less consistent initialization.

As reported in Tab.~\ref{tab:rolling_shutter}, our method consistently achieves lower \ac{ate}, \ac{are}, and reprojection error than Hyperion (Z-Splines) \cite{hyperion}.  
Under ideal conditions without distortion (global shutter), both methods perform comparably.  
However, when introducing a realistic rolling shutter readout of $1$\,ms, Hyperion frequently diverges for many pose blocks due to insufficient constraints and reduced scene overlap. The mean reprojection error can appear deceptively low in these cases, since reprojected scene landmarks falling outside the image plane are not counted by the metric.

\begin{figure}[t]
    \centering
    \includegraphics[width=\linewidth]{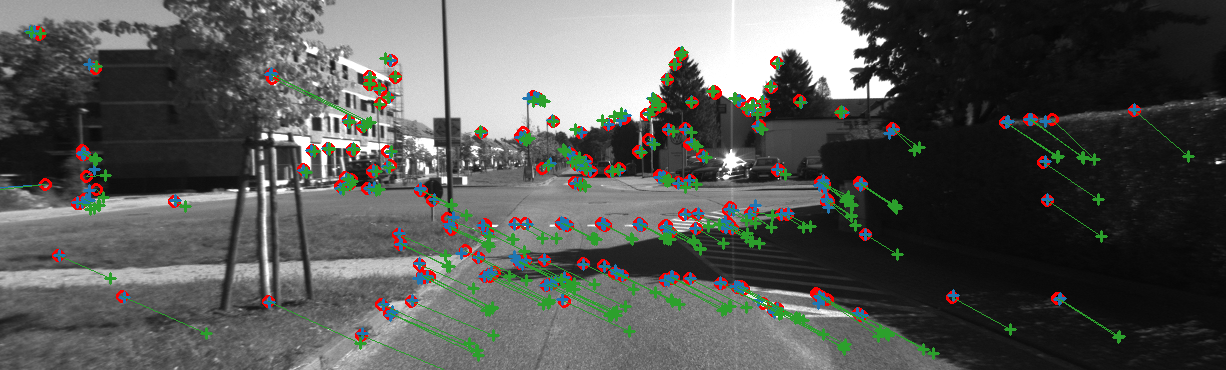}
    \caption{\textbf{Rolling shutter reprojection.} 
    Reprojection of optimized 3D scene landmarks on KITTI~\textit{06} \cite{geiger2013vision} under a $1$\,ms simulated readout time. 
    Hyperion \cite{hyperion} results are shown in \textcolor[rgb]{0.173,0.627,0.173}{green} and G-solver in \textcolor[rgb]{0.121,0.467,0.705}{blue}. 
    Nominal global shutter projections are shown in \textcolor[rgb]{1,0,0}{red}.  
    Our continuous-time \ac{gp} model explicitly accounts for row-wise exposure, producing sharper and more temporally consistent reprojections than the spline-based Hyperion approach.}
    \label{fig:rs_reprojection}
\end{figure}

In contrast, the \ac{gp} motion prior regularizes the solution even under extreme distortions, ensuring a well-conditioned optimization problem.  
G-solver produces smooth, temporally consistent trajectories and keeps the mean reprojection error below two pixels even at a readout time of $0.1$\,s.  
A qualitative comparison is shown in \figref{fig:rs_reprojection}, where the row-wise continuous-time modeling enables noticeably sharper and more accurate reprojections.

\begin{table}[h]
\centering
\scriptsize
\setlength{\tabcolsep}{2.5pt}
\begin{tabular}{@{}cc
cccccc@{}}
\toprule
\multicolumn{2}{c}{\multirow{2}{*}{KITTI}} & \multicolumn{5}{c}{Readout time [s]} \\ 
\cmidrule(lr){3-7}
 & & {0e+00} & {1e-04} & {1e-03} & {1e-02} & {1e-01} \\ 
\midrule

\multirow{3}{*}{Ours} 
 & ATE [m] & \textbf{0.671} & \textbf{0.676} & \textbf{0.676} & \textbf{0.684} & \textbf{0.726} \\
 & ARE [rad] & \textbf{0.007} & \textbf{0.007} & \textbf{0.008} & \textbf{0.007} & \textbf{0.008} \\
 & Mean Repr. Err. [pixel] & \textbf{1.704} & \textbf{1.708} & \textbf{1.709} & \textbf{1.715} & \textbf{1.844} \\ 
\midrule

\multirow{3}{*}{Z-Spline} 
 & ATE [m] & 0.692 & 94.510 & 1963.040 & 200.168 & 733.028 \\
 & ARE [rad] & \textbf{0.007} & 0.194 & 0.508 & 0.350 & 1.577 \\
 & Mean Repr. Err. [pixel] & 1.717 & 2.099 & 1.884 & 2.301 & 8.142 \\

\bottomrule
\end{tabular}
\caption{\textbf{Rolling shutter readout performance.} 
Each row reports \ac{ate}, \ac{are}, and mean reprojection error for a given method, and each column corresponds to a different simulated readout time. 
\ac{ate} and \ac{are} are computed by matching each ground truth pose with the method’s continuous-time pose interpolated at the same timestamp. 
Mean reprojection error is evaluated only for scene landmarks whose projections fall within the image plane.}
\label{tab:rolling_shutter}
\end{table}

\begin{figure}[t]
    \centering
    \includegraphics[width=\linewidth]{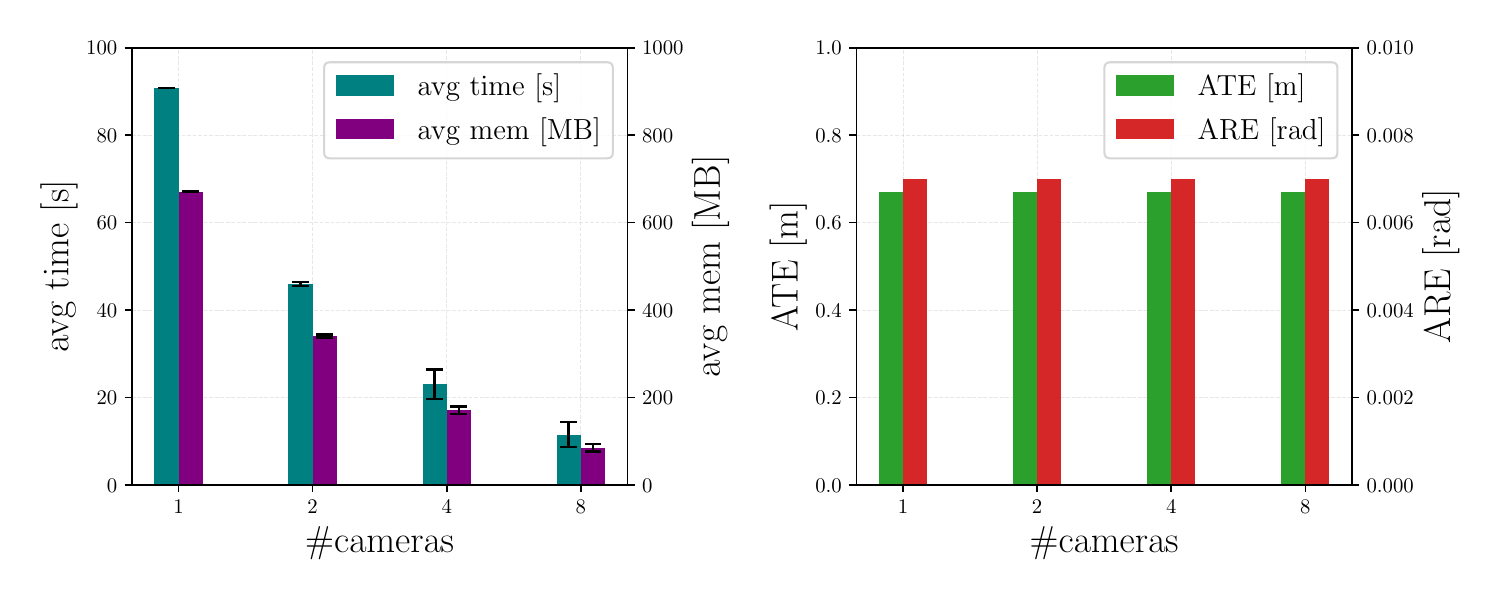}
    \caption{\textbf{Multi-camera optimization scalability.} 
    (Left) The average computation time and memory per camera graph decrease as the number of camera segments increases, as the optimization load is distributed across multiple independent subgraphs. 
    (Right) Estimation accuracy remains stable (\ac{ate} and \ac{are} stay nearly constant), showing that distributing the optimization lowers computational cost without degrading accuracy.}
    \label{fig:multi_camera}
\end{figure}

\paragraph{Multi-Camera BA}
For the multi-camera experiment, ORB-SLAM3 \cite{campos2021orb} is used as the front-end to extract stereo measurements. 
The KITTI \textit{06} \cite{geiger2013vision} sequence is partitioned into multiple temporal segments, each treated as an independent camera stream with its own local factor graph. To enable joint optimization across cameras, we introduce equality factors that link corresponding scene landmarks observed from different viewpoints. 
These constraints are established during a lightweight handshake before optimization, enabling fully distributed inference.
Consistent with the previous experiment, \ac{gp} hyperparameters are initialized from KITTI~\textit{05}.
As shown in \figref{fig:multi_camera}, distributing the optimization significantly reduces average runtime and memory per camera. Importantly, the final reconstructed trajectories in both distributed and centralized configurations exhibit similar \ac{ate} and \ac{are}, indicating that parallelism can be leveraged without compromising accuracy.

\section{Conclusion}
\label{sec:conclusion}
We presented G-solver, a fully Gaussian and distributed framework for
continuous-time SLAM that combines Gaussian Process motion priors with Gaussian
Belief Propagation. Our formulation enables uncertainty-aware trajectory
estimation, scalable message-passing inference, and consistent handling of
asynchronous measurements within a unified continuous-time model. Through
extensive experiments, including rolling shutter and multi-camera distributed
optimization, we showed that G-solver consistently improves accuracy and
robustness while matching the cost of state-of-the-art spline-based and
centralized methods, without engineering overhead. Looking forward, we plan to
develop more compact message-passing schedules to reduce communication and
runtime, and to incorporate richer \ac{gp} priors, such as constant-acceleration
or jerk-driven models, for higher-fidelity motion representation in challenging
dynamic scenarios.

{   
    \small
    \bibliographystyle{ieeetr}
    \bibliography{main}
}

\newpage
\section{Supplementary Material}
\label{sec:supplementary}

This section provides additional details to complement the main paper. We include a complete description of the inference components used in our framework, extended qualitative results, and additional visualizations such as continuous-time trajectory plots and rolling-shutter reprojection examples. These materials are intended to make the work fully reproducible and to offer further insight beyond the quantitative results presented in the main text.

\subsection{Gaussian Belief Propagation on Manifold}
\ac{gbp} is an iterative optimization algorithm that operates in a fully distributed manner via message passing. It is a special case of \ac{bp} in which the variables are normally distributed. The notation from \cite{Bishop2006, Davison2019} is followed throughout. In the remainder of this section, the inference problem is assumed to be expressed as a factor graph.
Each node in a factor graph can either be a variable $x_i$:
\begin{equation}
    x_i \sim \cN(\bmu_{x_i}, \bSigma_{x_i}) = \cN^{-1}(\bfeta_{x_i}, \bLambda_{x_i}),
\end{equation}
or a factor $f_j$:
\begin{equation}
    f_j \sim \cN^{-1}(\bfeta_{f_j}, \bLambda_{f_j}),
\end{equation}
where the precision matrix is the inverse of the covariance matrix ($\bLambda = \bSigma^{-1}$) and the information vector satisfies the relation $\bfeta = \bLambda \bmu$.
The goal of \ac{gbp} is to find the configuration of variables that minimizes the residual induced by each factor, as follows:
\begin{equation}
    \argmin_{k \in n(f_j)} \lVert f_j(x_k) \rVert^2_{\bLambda_k}.
\end{equation}
$x_k \subset n(f_j)$ is the subset of variables from which the measurement encoded in $f_j$ depends.
\ac{gbp} is a decentralized iterative algorithm consisting of four main steps: variable update, variable-to-factor message generation, factor update, and factor-to-variable message generation. Both factor and variable nodes perform computations locally and communicate with their neighbors through messages that encode Gaussian distributions.
Two types of messages are distinguished: the variable-to-factor message, $m_{x_i\rightarrow f_j}$, and the factor-to-variable message, $m_{f_j\rightarrow x_i}$. In the following, the canonical or moment parameterization of a Gaussian is used when convenient, while the footers indicate the direction of the message.
\\To include manifolds each variable or message distribution is parametrized with a mean $\bmu \in \Manifold$ on the manifold and a covariance matrix $\bLambda^{-1} \in \bbR^{\mathrm{dim}(\Tangent{\bmu}) \times \mathrm{dim}(\Tangent{\bmu})}$ in the tangent space at the mean.
\\Each time a message parametrized in this manner is received, it must be internally remapped to the Euclidean tangent. This step is necessary to perform operations on the underlying Gaussian distribution. Let $m_{in} \sim \cN(\bmu_{in}, \bLambda_{in}^{-1})$ denote a generic incoming message. The corresponding parameters in the tangent space are given by $m_{in}^{\btau} \sim \cN(\btau_{in}, (\bLambda_{in}^{\btau})^{-1})$. Using the $\boxplus$ notation, this can be expressed as:
\begin{align}
    \btau_{in} &= \bmu_{in} \boxminus \bmu_{0}, \\
    \bLambda_{in}^{\btau} &\approx \bLambda_{in} \label{eq:lambdain_supp}.
\end{align}
Here $\bmu_0$ is the most recent estimate of the mean, at the origin of the tangent space.
\\Similarly, whenever an outward message must be sent, its parameters are remapped from the tangent space back to the manifold, as follows:
\begin{align}
    \bmu_{out} &= \bmu_0 \boxplus \btau_{out}, \\
    \bLambda_{out} &\approx \bLambda_{out}^{\btau} \label{eq:lambdaout}.
\end{align}
In the equation, $m_{out}^{\btau} \sim \cN(\btau_{out}, (\bLambda_{out}^{\btau})^{-1})$ represents the outward message in the tangent space, while $m_{out} \sim \cN(\bmu_{out}, \bLambda_{out}^{-1})$ denotes its manifold counterpart. \eqref{eq:lambdain} and \eqref{eq:lambdaout} hold for the small perturbations computed by the iterative \ac{gbp} scheme. Using these procedures to map estimates forward and backward between the tangent and manifold spaces provides a straightforward extension of \ac{gbp} to non-Euclidean domains.

\paragraph{Variable Update}
\label{sec:variable_update}
The variable update step recomputes the new estimate of a variable given all incoming messages from the connected factors. Each message encodes the probability distribution of the receiving variable as suggested by the sending factor. With all messages expressed in the same tangent space, $m_{f_k \rightarrow x_i}^{\btau} \sim \cN(\btau_{f_k \rightarrow x_i}, (\bLambda_{f_k \rightarrow x_i}^{\btau})^{-1})$, the updated distribution can be determined as:
\begin{align}
    \bLambda_{new}^{\btau} &= \sum_{k \in n(x_i)} \bLambda_{f_k \rightarrow x_i}^{\btau}, \\
    \btau_{new} &= (\bLambda_{new}^{\btau})^{-1} \sum_{k \in n(x_i)} \bLambda_{f_k \rightarrow x_i}^{\btau} \btau_{f_k \rightarrow x_i},
\end{align}
where $f_k \subset n(x_i)$ is the subset of factors in which the variable is involved. Finally, by projecting the new estimate back to the manifold, the updated estimate of the variable is obtained as $x_i \sim \cN(\bmu_{new}, \bLambda_{new}^{-1})$.

\paragraph{Variable-to-Factor Message}
\label{sec:v2f_message}
A variable-to-factor message encodes the distribution of the sending variable that the receiving factor will use to compute the cost function, and thus to perform optimization. The generation of such a message is identical to the variable update, except that all incoming messages are considered except for the one arriving from the recipient factor $f_r$. Defining the incoming messages as $m_{f_k \rightarrow x_i}^{\btau} \sim \cN(\btau_{f_k \rightarrow x_i}, (\bLambda_{f_k \rightarrow x_i}^{\btau})^{-1})$, already expressed in the tangent space, the updated distribution can be determined as:
\begin{align}
    \bLambda_{new}^{\btau} &= \sum_{k \in n(x_i) \setminus f_r} \bLambda_{f_k \rightarrow x_i}^{\btau}, \\
    \btau_{new} &= (\bLambda_{new}^{\btau})^{-1} \sum_{k \in n(x_i) \setminus f_r} \bLambda_{f_k \rightarrow x_i}^{\btau} \btau_{f_k \rightarrow x_i}.
\end{align}
Finally, by projecting the perturbation back to the manifold, the outgoing message is obtained as $m_{x_i \rightarrow f_r} \sim \cN(\bmu_{x_i \rightarrow f_r}, \bLambda_{x_i \rightarrow f_r}^{-1})$.

\paragraph{Factor Update}
\label{sec:factor_update}
This step is crucial to keep the factor updated with respect to the estimates of the connected variables. As new variable-to-factor messages arrive at the factor node, the cost function must be recomputed. Given an error function $\be(x_k)$, a factor function is defined as:
\begin{align}
    f_j(x_k) &= K e^{-E_j(x_k)}, \\
    E_j(x_k) &= \frac{1}{2} \lVert \be_j(x_k)) \rVert^2_{\bLambda_j}.
    \label{eq:energy}
\end{align}
Here, $x_k \subset n(f_j)$ is the set of variables that appear in the factor. The factor used in the \ac{gbp} algorithm must be computed in the Euclidean tangent space and expressed in canonical form. A common choice is to center the tangent space at the most recent mean $\bmu_0$ of the variables $x_0$. Defining the error Jacobian as $\bJ = \frac{\partial \be_j(\bmu_k)}{\partial \bmu}|_{\bmu = \bmu_0}$, the following formulas are obtained:
\begin{align}
    \bfeta_{new} &= \bJ^\top \bLambda_j (-\be_j(x_0)),\\
    \bLambda_{new} &= \bJ^\top \bLambda_j \bJ.
\end{align}
This provides the updated estimate of the factor, expressed as $f_j \sim \cN^{-1}(\bfeta_{new}, \bLambda_{new})$.

\paragraph{Factor-to-Variable Message}
Finally, the factor-to-variable message encodes the distribution of the receiving variable as estimated by the sending factor through the minimization of the cost function. The computation process is described in detail in the main paper.
    
\subsection{Qualitative Results}
\begin{figure}[ht!]
\vspace{0.85cm}
    \centering

    \begin{subfigure}{0.40\textwidth}
        \centering
        \includegraphics[width=\textwidth]{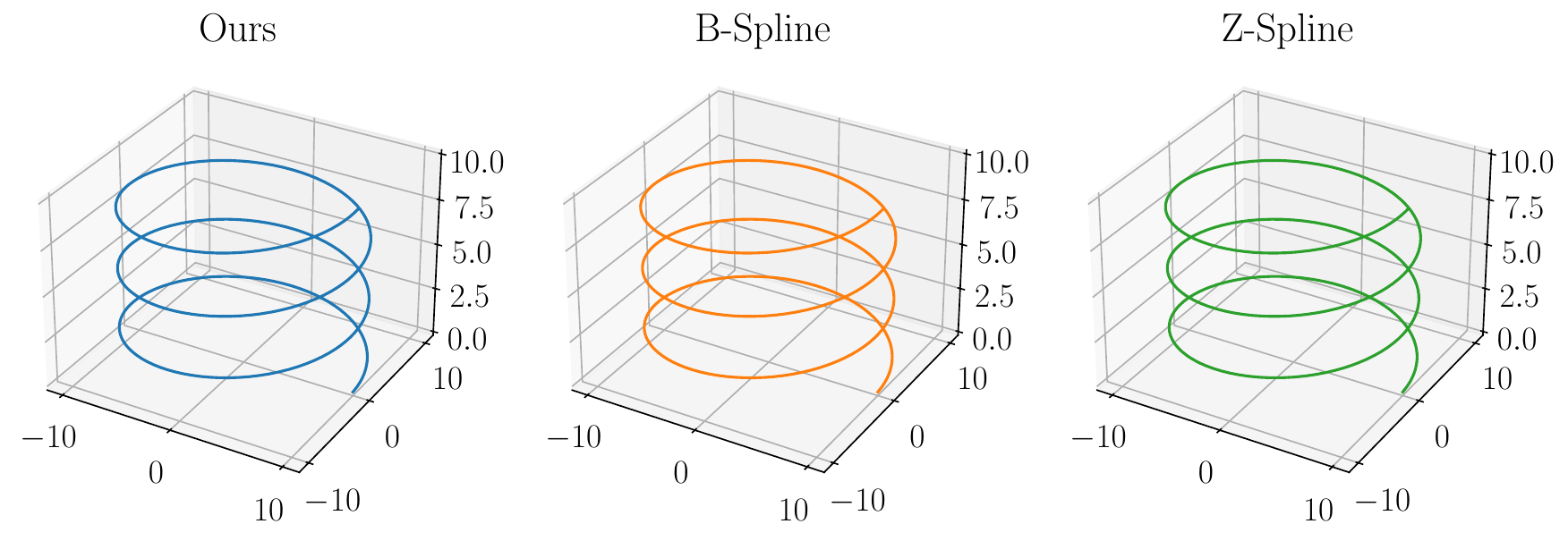}
        \caption{$\eta_{ig}=1, \sigma=10^{-4}$ (both $\mathrm{[m]}$ and $\mathrm{[rad]}$).}
    \end{subfigure}
    \vspace{0.2cm}

    \begin{subfigure}{0.40\textwidth}
        \centering
        \includegraphics[width=\textwidth]{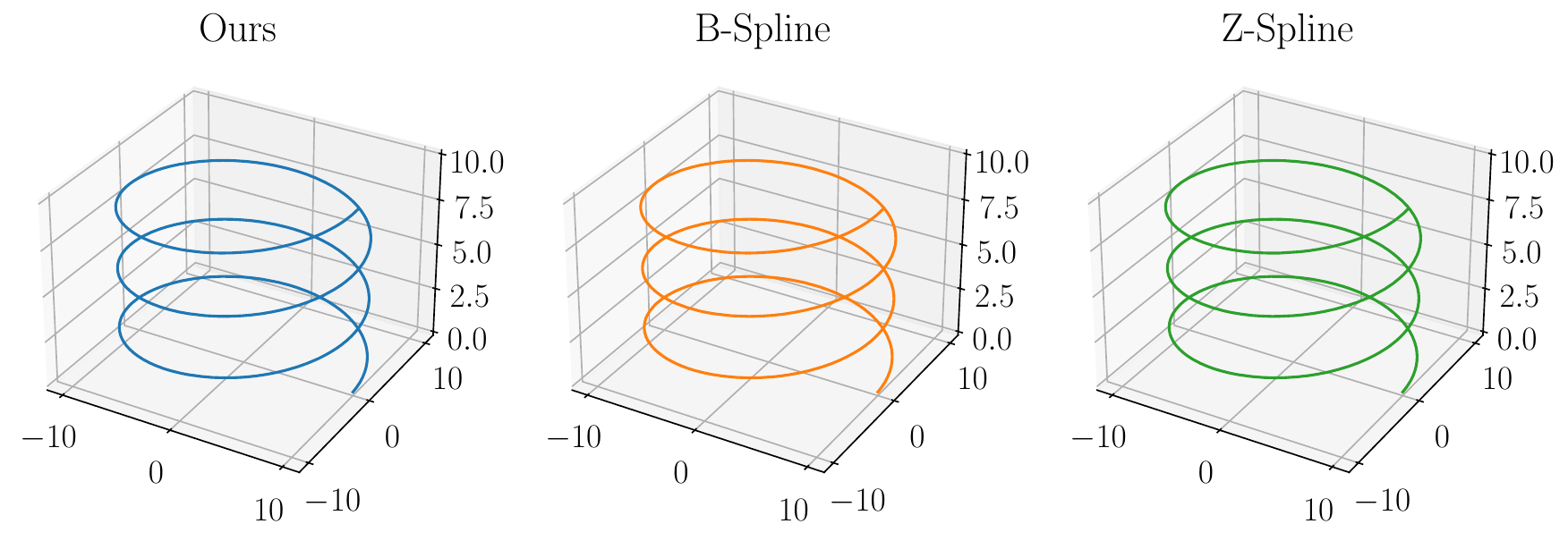}
        \caption{$\eta_{ig}=1, \sigma=10^{-3}$ (both $\mathrm{[m]}$ and $\mathrm{[rad]}$).}
    \end{subfigure}
    \vspace{0.2cm}

    \begin{subfigure}{0.40\textwidth}
        \centering
        \includegraphics[width=\textwidth]{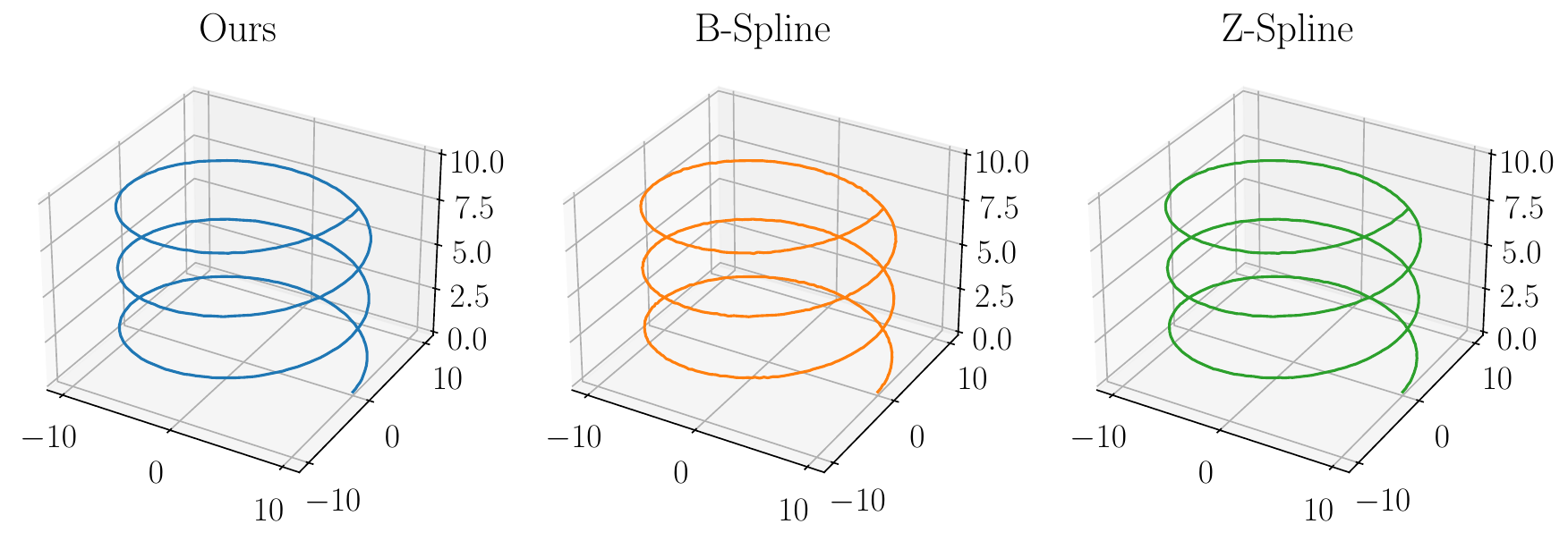}
        \caption{$\eta_{ig}=1, \sigma=10^{-2}$ (both $\mathrm{[m]}$ and $\mathrm{[rad]}$).}
    \end{subfigure}
    \vspace{0.2cm}

    \begin{subfigure}{0.40\textwidth}
        \centering
        \includegraphics[width=\textwidth]{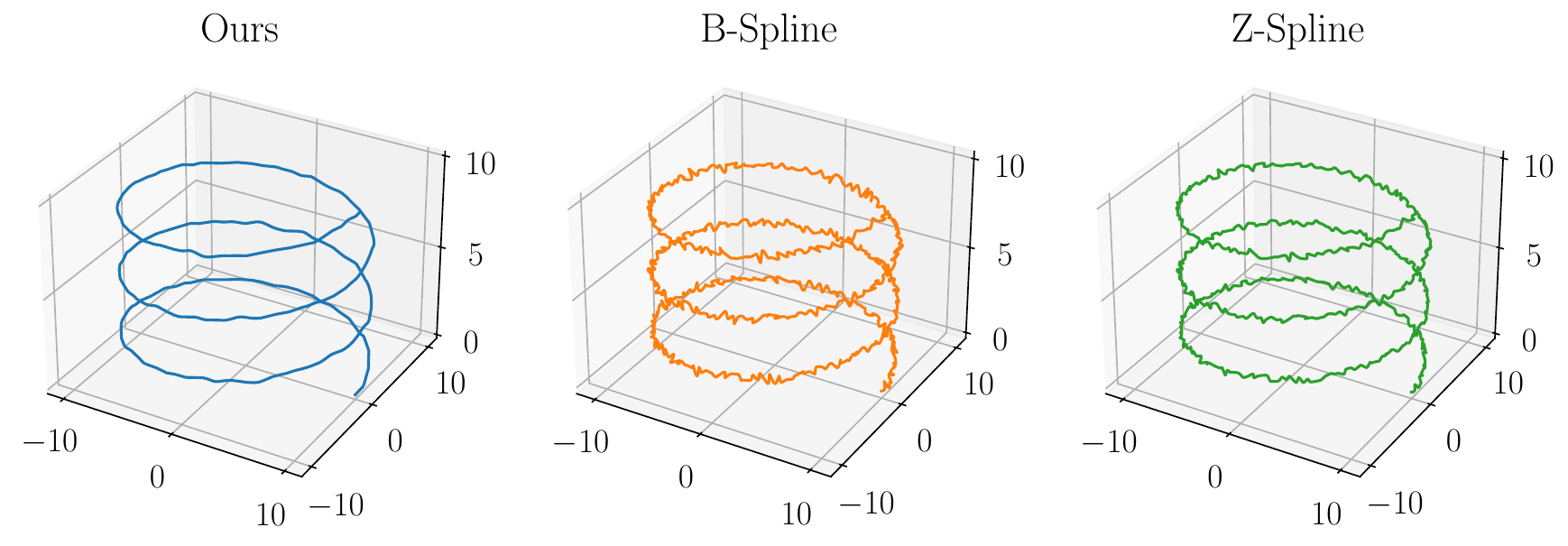}
        \caption{$\eta_{ig}=1, \sigma=10^{-1}$ (both $\mathrm{[m]}$ and $\mathrm{[rad]}$).}
    \end{subfigure}
    \vspace{0.2cm}

    \begin{subfigure}{0.40\textwidth}
        \centering
        \includegraphics[width=\textwidth]{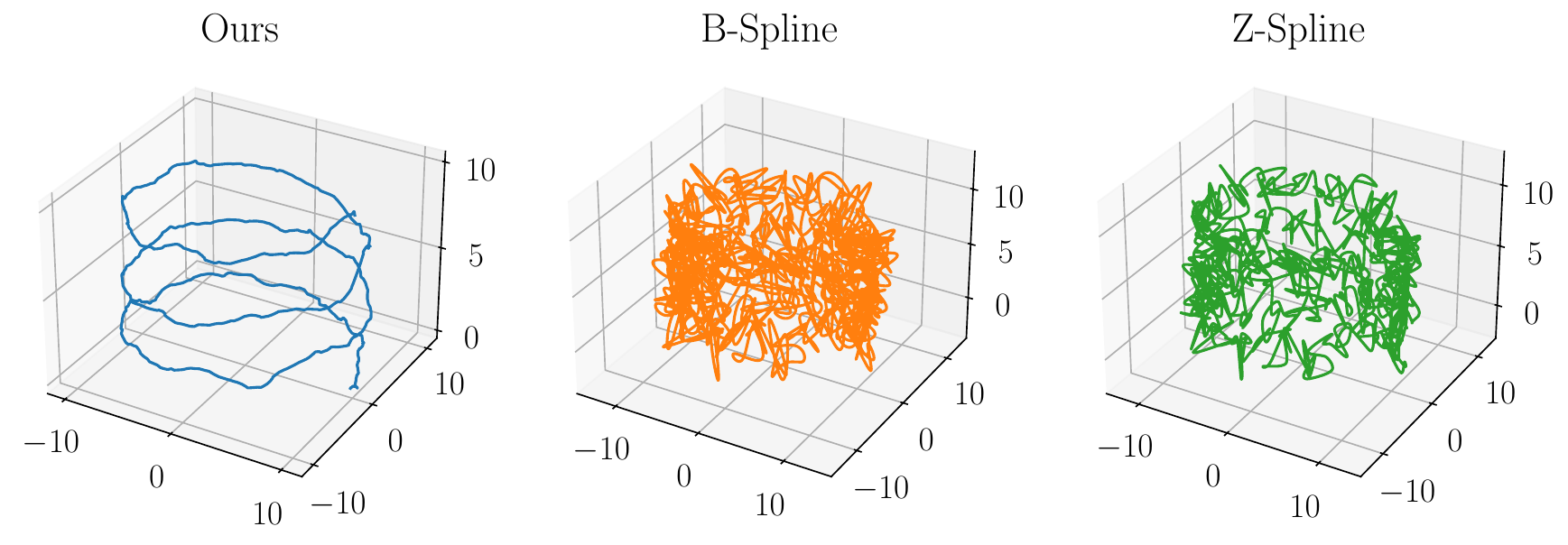}
        \caption{$\eta_{ig}=1, \sigma=1$ (both $\mathrm{[m]}$ and $\mathrm{[rad]}$).}
    \end{subfigure}
    \vspace{0.2cm}

    \begin{subfigure}{0.40\textwidth}
        \centering
        \includegraphics[width=\textwidth]{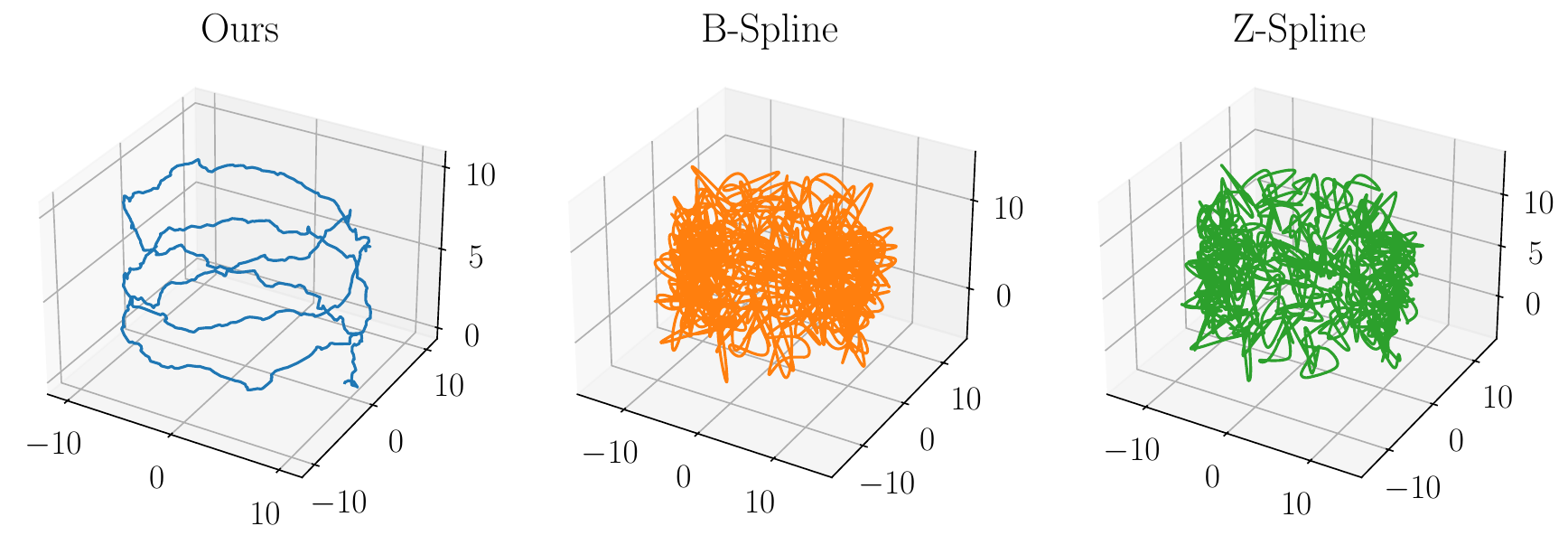}
        \caption{$\eta_{ig}=1, \sigma=1.5$ (both $\mathrm{[m]}$ and $\mathrm{[rad]}$).}
    \end{subfigure}

    \caption{\textbf{Prior-based experiments.} Comparison between G-solver and Hyperion using the \textit{helix} trajectory under varying measurement Gaussian noise standard deviations $\sigma$. The initial guess is perturbed with a standard deviation of $\eta_{ig} = 1$ (applied to both positions $\mathrm{[m]}$ and orientations $\mathrm{[rad]}$). All methods query the trajectory at a resolution $100\times$ finer than the control-point spacing.}
    \label{fig:prior1}
\end{figure}

\begin{figure}[ht!]
    \centering

    \begin{subfigure}{0.40\textwidth}
        \centering
        \includegraphics[width=\textwidth]{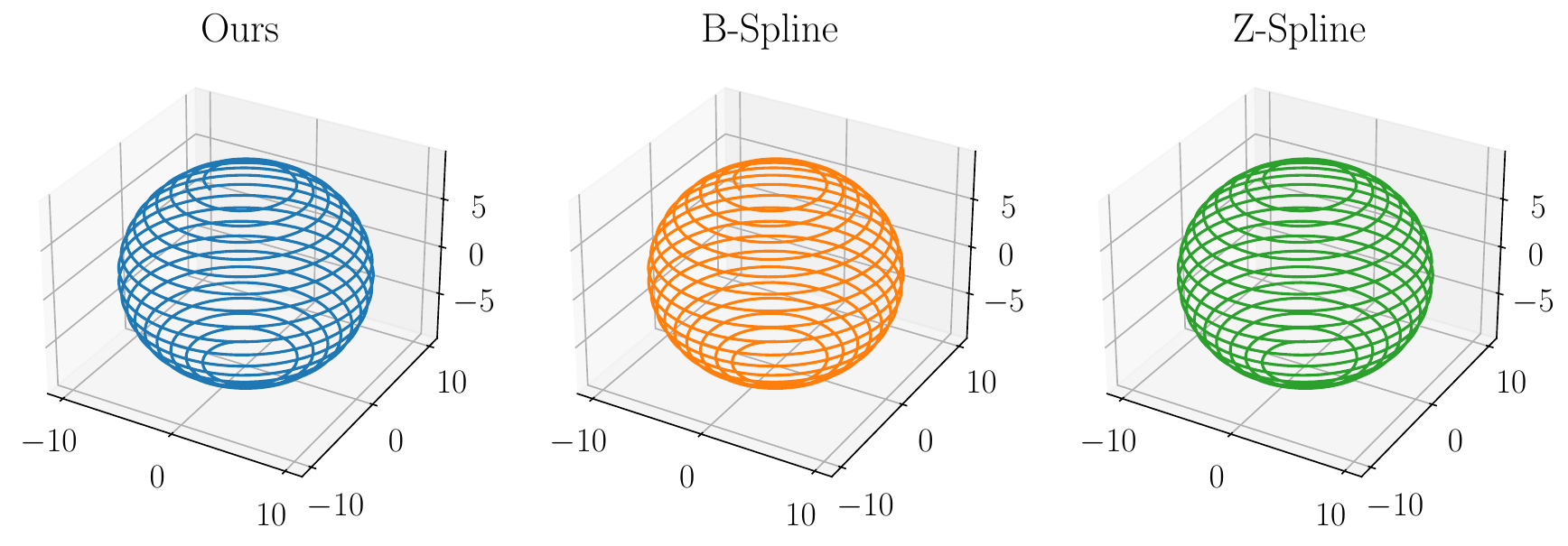}
        \caption{$\eta_{ig}=1, \sigma=10^{-4}$ (both $\mathrm{[m]}$ and $\mathrm{[rad]}$).}
    \end{subfigure}
    \vspace{0.2cm}

    \begin{subfigure}{0.40\textwidth}
        \centering
        \includegraphics[width=\textwidth]{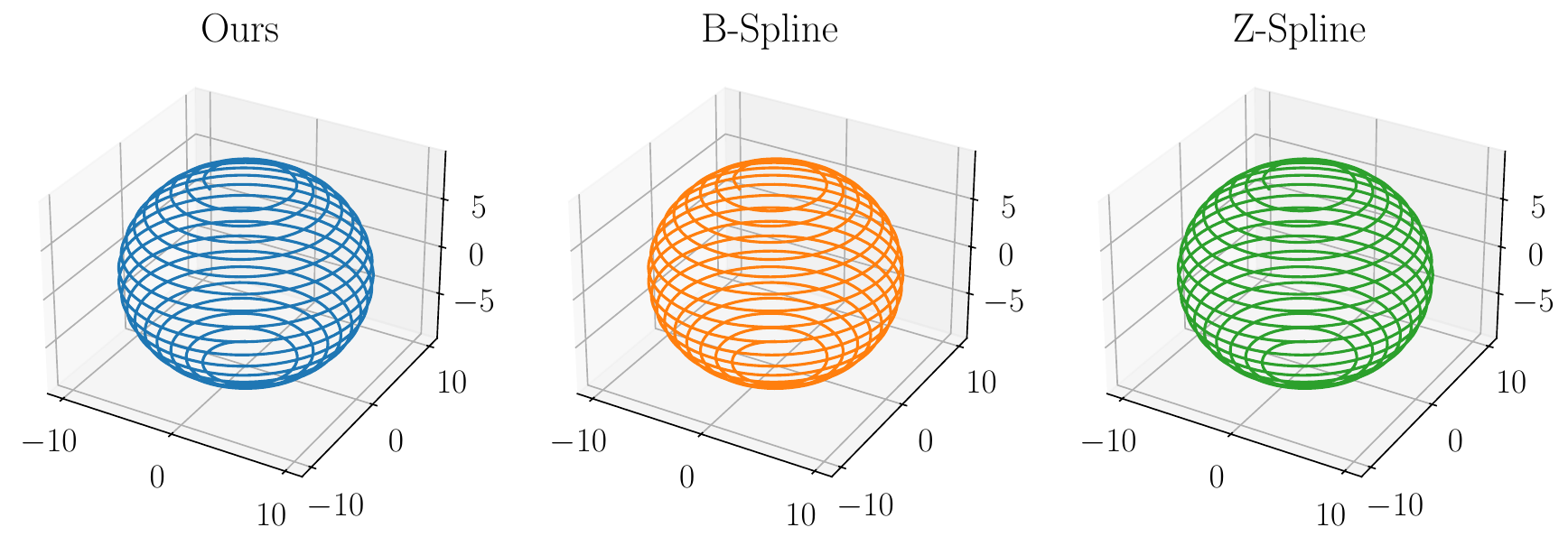}
        \caption{$\eta_{ig}=1, \sigma=10^{-3}$ (both $\mathrm{[m]}$ and $\mathrm{[rad]}$).}
    \end{subfigure}
    \vspace{0.2cm}

    \begin{subfigure}{0.40\textwidth}
        \centering
        \includegraphics[width=\textwidth]{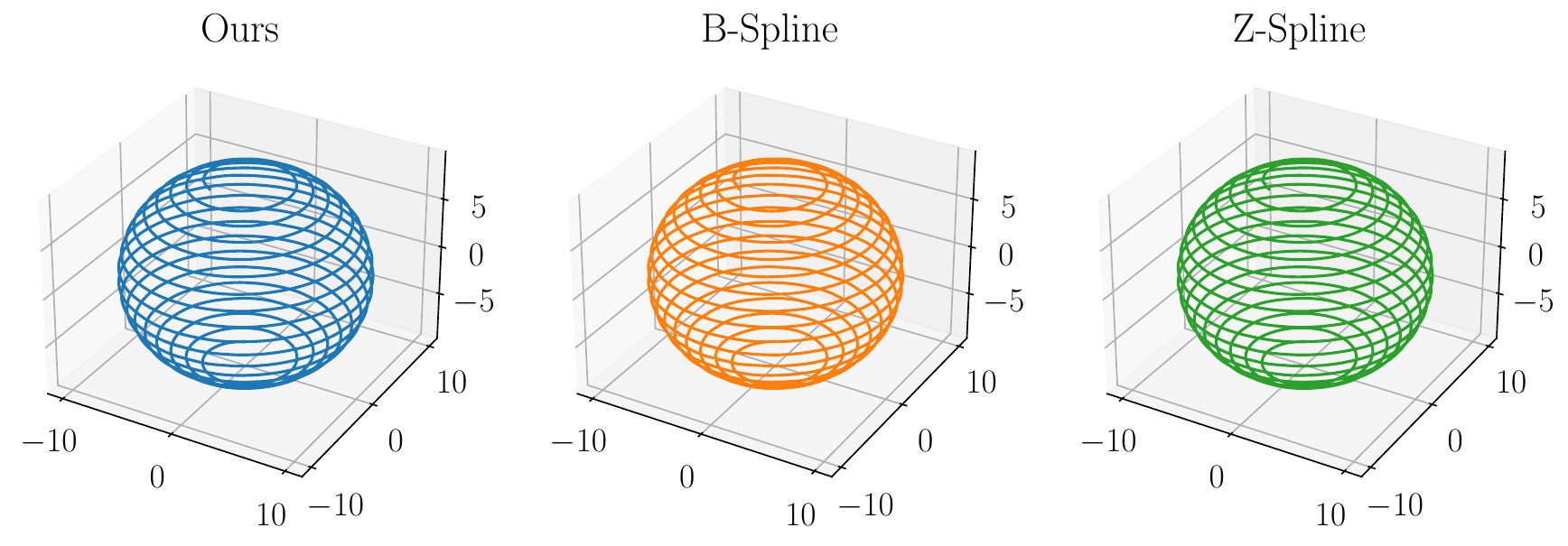}
        \caption{$\eta_{ig}=1, \sigma=10^{-2}$ (both $\mathrm{[m]}$ and $\mathrm{[rad]}$).}
    \end{subfigure}
    \vspace{0.2cm}

    \begin{subfigure}{0.40\textwidth}
        \centering
        \includegraphics[width=\textwidth]{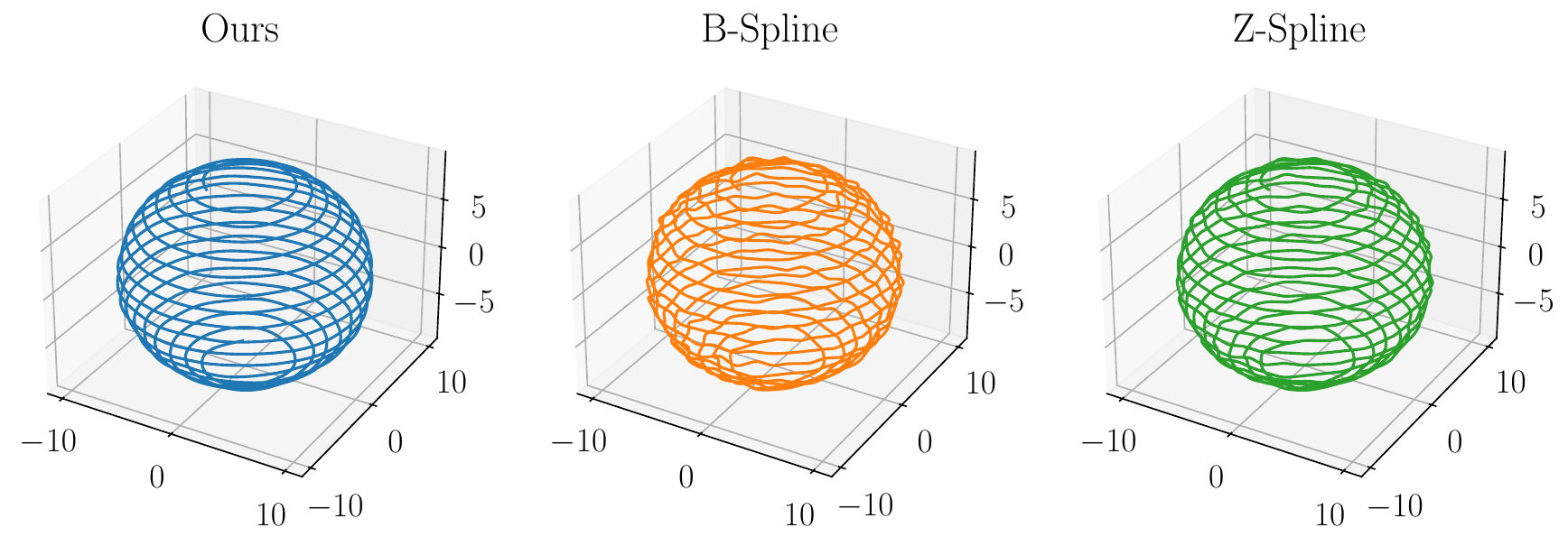}
        \caption{$\eta_{ig}=1, \sigma=10^{-1}$ (both $\mathrm{[m]}$ and $\mathrm{[rad]}$).}
    \end{subfigure}
    \vspace{0.2cm}

    \begin{subfigure}{0.40\textwidth}
        \centering
        \includegraphics[width=\textwidth]{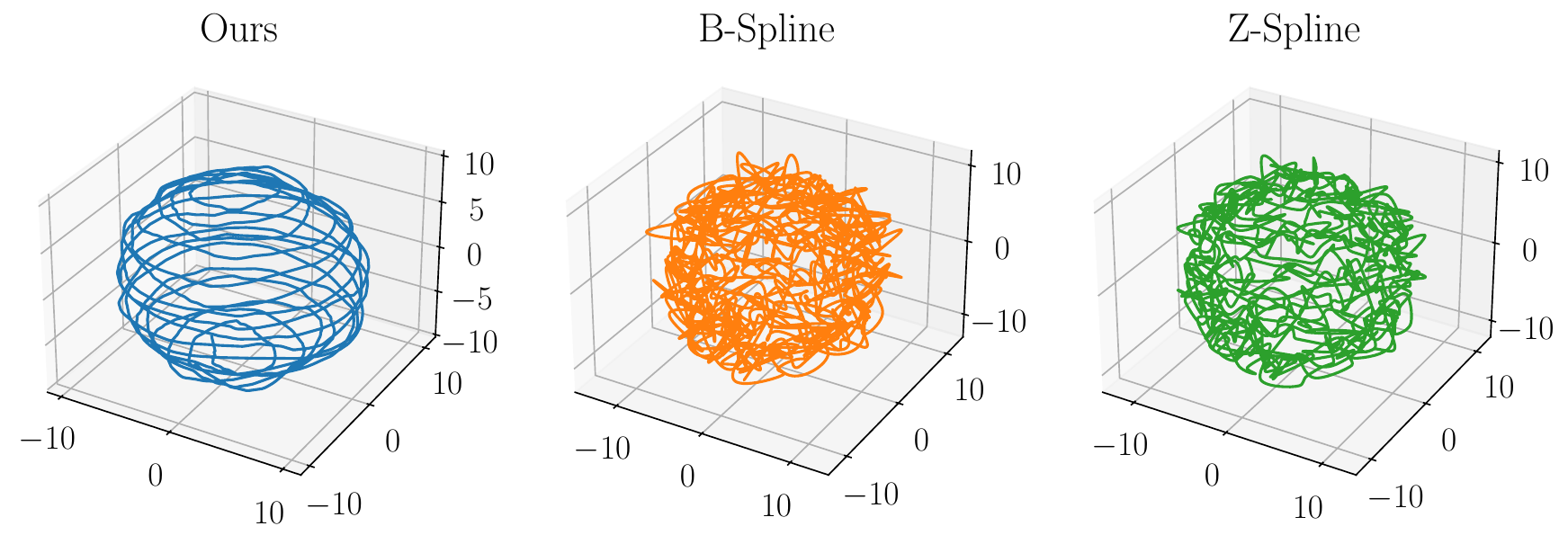}
        \caption{$\eta_{ig}=1, \sigma=1$ (both $\mathrm{[m]}$ and $\mathrm{[rad]}$).}
    \end{subfigure}
    \vspace{0.2cm}

    \begin{subfigure}{0.40\textwidth}
        \centering
        \includegraphics[width=\textwidth]{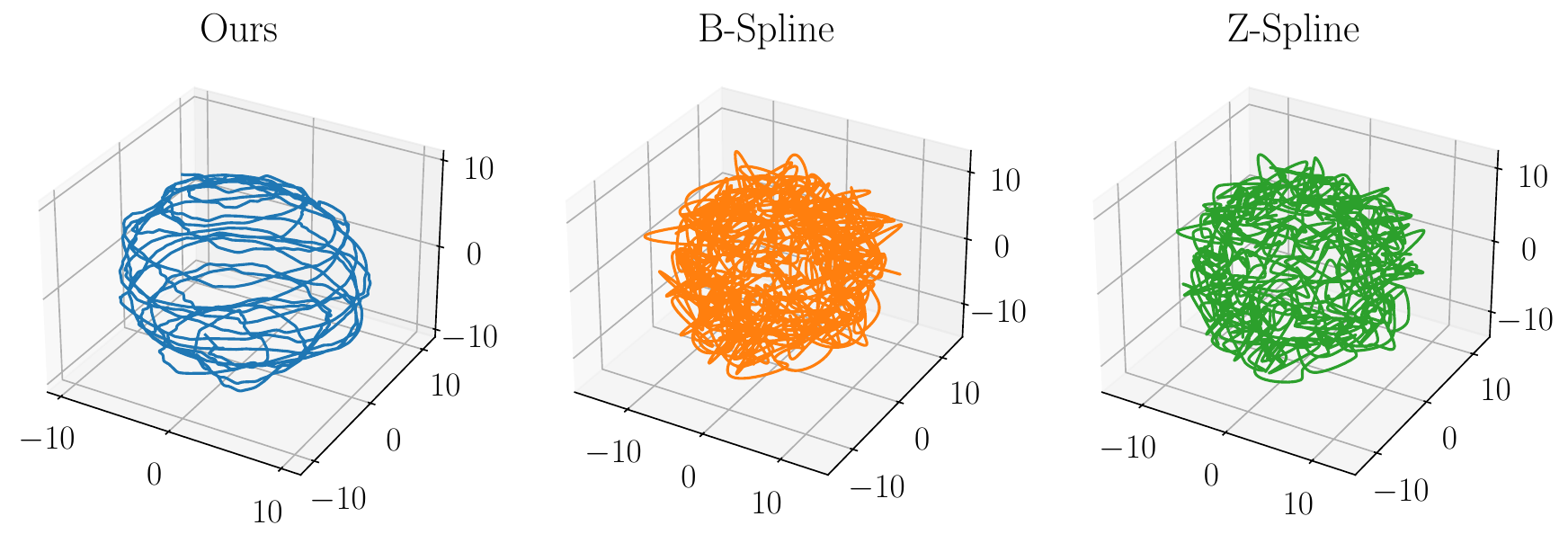}
        \caption{$\eta_{ig}=1, \sigma=1.5$ (both $\mathrm{[m]}$ and $\mathrm{[rad]}$).}
    \end{subfigure}

    \caption{\textbf{Prior-based experiments.} Comparison between G-solver and Hyperion using the \textit{sphere} trajectory under varying measurement Gaussian noise standard deviations $\sigma$. The initial guess is perturbed with a standard deviation of $\eta_{ig} = 1$ (applied to both positions $\mathrm{[m]}$ and orientations $\mathrm{[rad]}$). All methods query the trajectory at a resolution $100\times$ finer than the control-point spacing.}
    \label{fig:prior2}
\end{figure}

\begin{figure}[ht!]
    \centering

    \begin{subfigure}{0.40\textwidth}
        \centering
        \includegraphics[width=\textwidth]{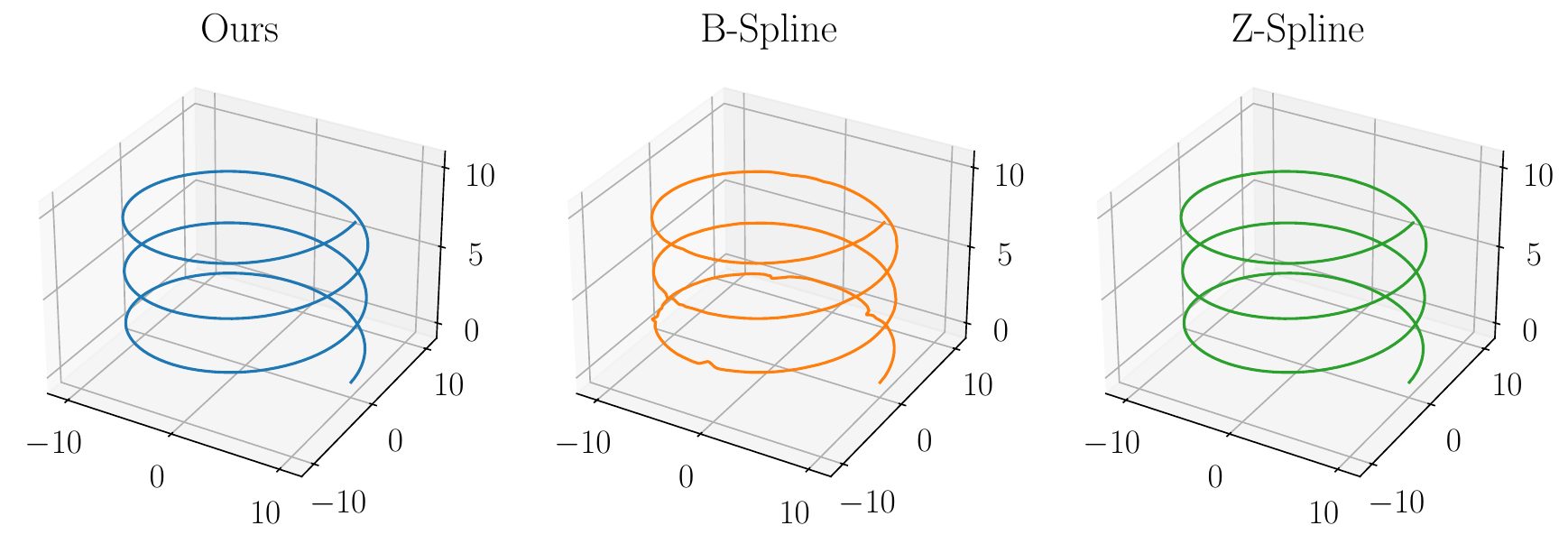}
        \caption{$\eta_{ig}=1, \sigma=10^{-4}$ (both $\mathrm{[m]}$ and $\mathrm{[rad]}$).}
    \end{subfigure}
    \vspace{0.2cm}

    \begin{subfigure}{0.40\textwidth}
        \centering
        \includegraphics[width=\textwidth]{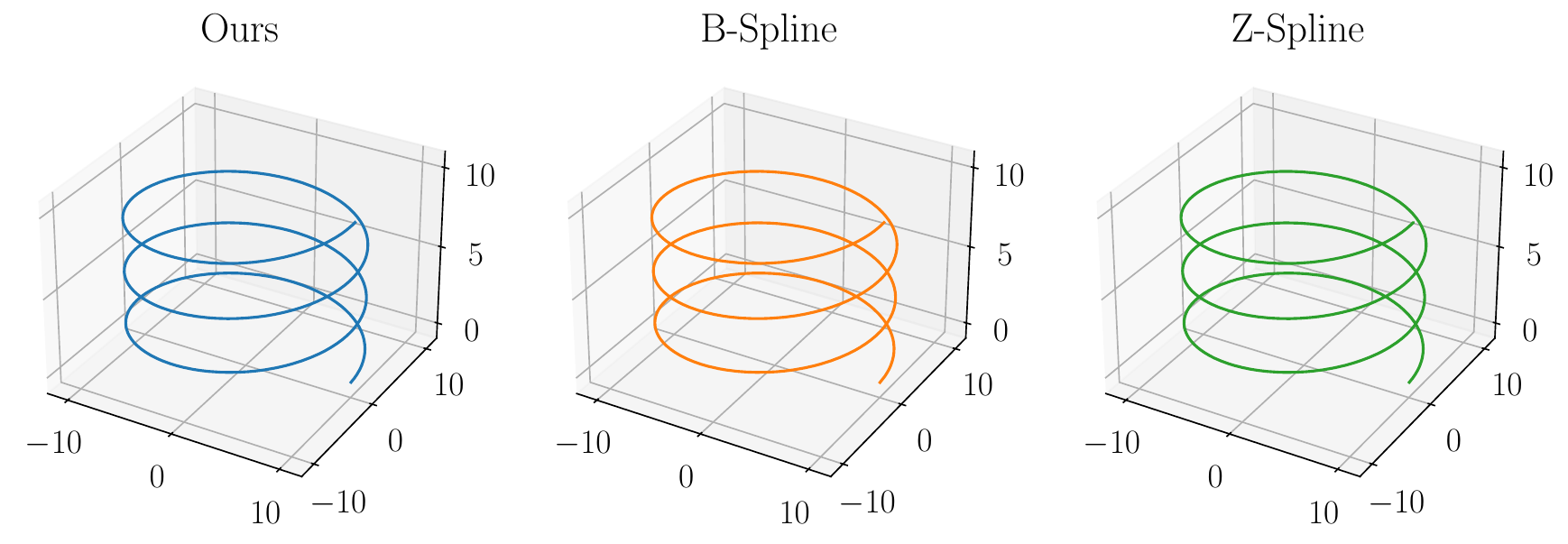}
        \caption{$\eta_{ig}=1, \sigma=10^{-3}$ (both $\mathrm{[m]}$ and $\mathrm{[rad]}$).}
    \end{subfigure}
    \vspace{0.2cm}

    \begin{subfigure}{0.40\textwidth}
        \centering
        \includegraphics[width=\textwidth]{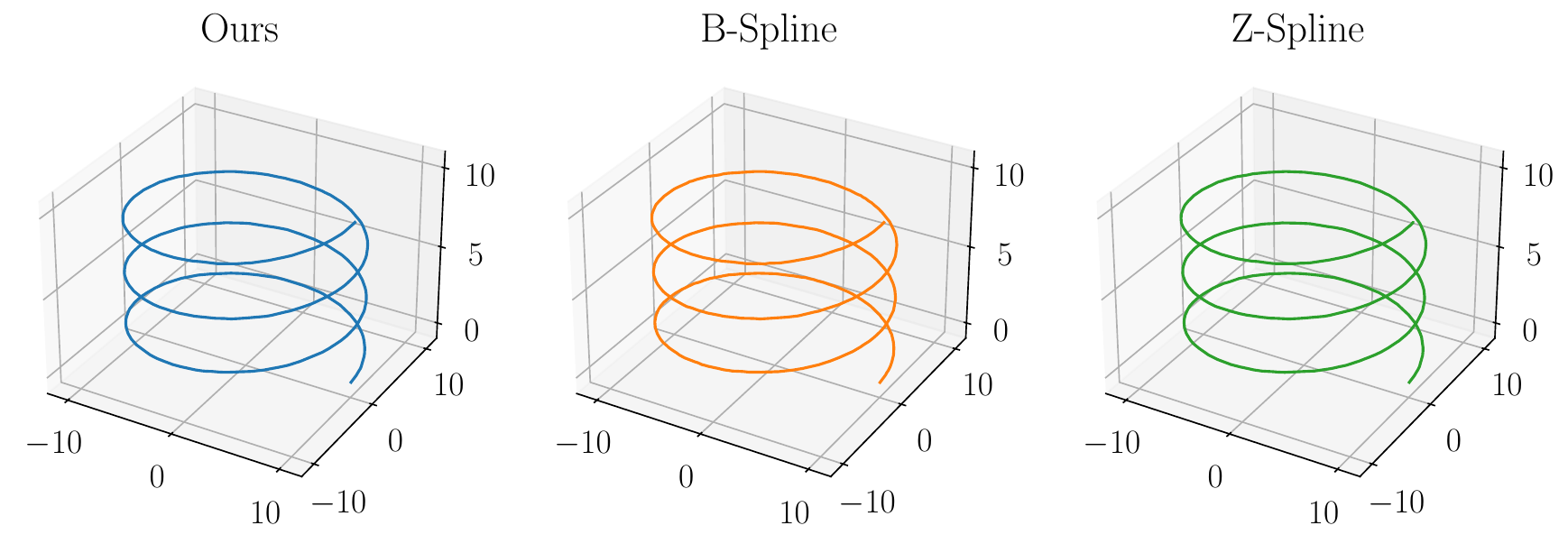}
        \caption{$\eta_{ig}=1, \sigma=10^{-2}$ (both $\mathrm{[m]}$ and $\mathrm{[rad]}$).}
    \end{subfigure}
    \vspace{0.2cm}

    \begin{subfigure}{0.40\textwidth}
        \centering
        \includegraphics[width=\textwidth]{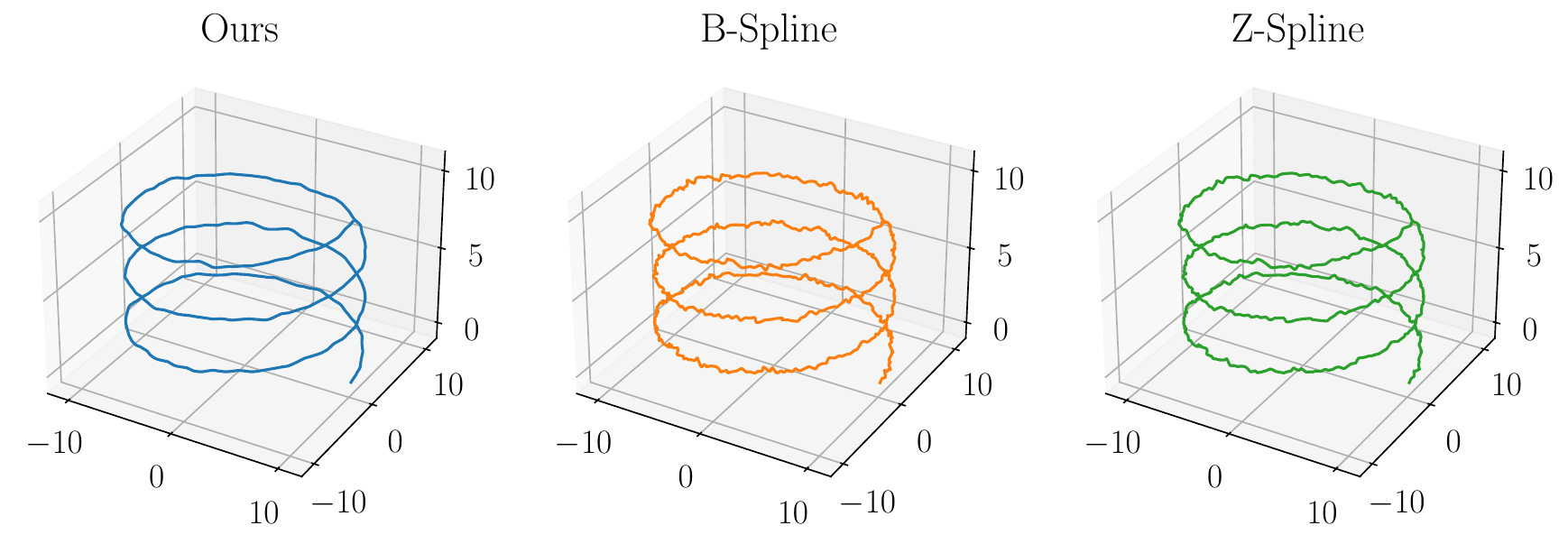}
        \caption{$\eta_{ig}=1, \sigma=10^{-1}$ (both $\mathrm{[m]}$ and $\mathrm{[rad]}$).}
    \end{subfigure}
    \vspace{0.2cm}

    \begin{subfigure}{0.40\textwidth}
        \centering
        \includegraphics[width=\textwidth]{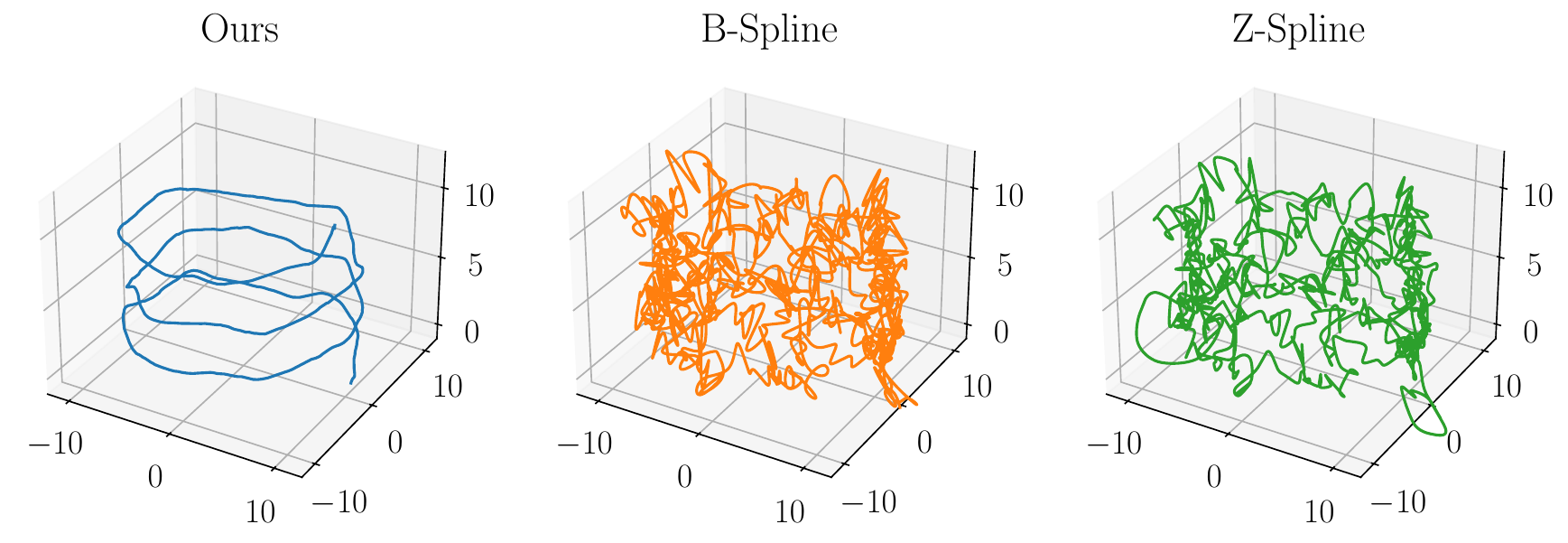}
        \caption{$\eta_{ig}=1, \sigma=1$ (both $\mathrm{[m]}$ and $\mathrm{[rad]}$).}
    \end{subfigure}
    \vspace{0.2cm}

    \begin{subfigure}{0.40\textwidth}
        \centering
        \includegraphics[width=\textwidth]{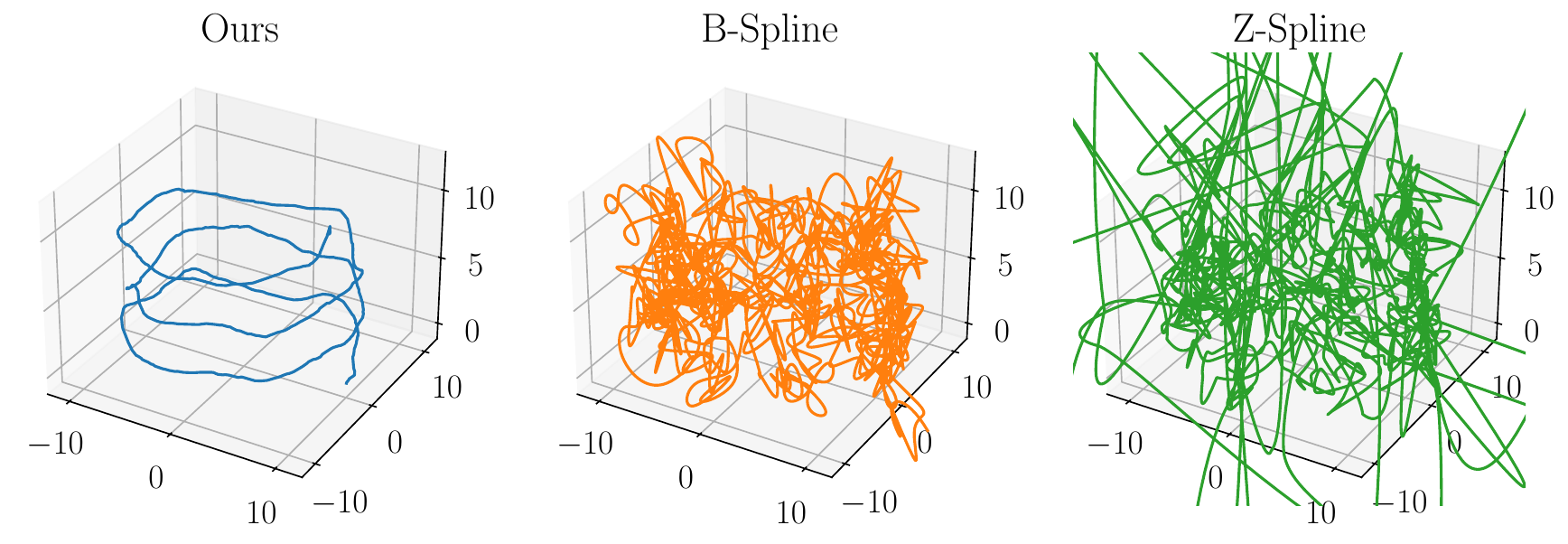}
        \caption{$\eta_{ig}=1, \sigma=1.5$ (both $\mathrm{[m]}$ and $\mathrm{[rad]}$).}
    \end{subfigure}

    \caption{\textbf{PGO experiments.} Comparison between G-solver and Hyperion using the \textit{helix} trajectory under varying measurement Gaussian noise standard deviations $\sigma$. The initial guess is perturbed with a standard deviation of $\eta_{ig} = 1$ (applied to both positions $\mathrm{[m]}$ and orientations $\mathrm{[rad]}$). All methods query the trajectory at a resolution $100\times$ finer than the control-point spacing.}
    \label{fig:pgo1}
\end{figure}

\begin{figure}[ht!]
    \centering

    \begin{subfigure}{0.40\textwidth}
        \centering
        \includegraphics[width=\textwidth]{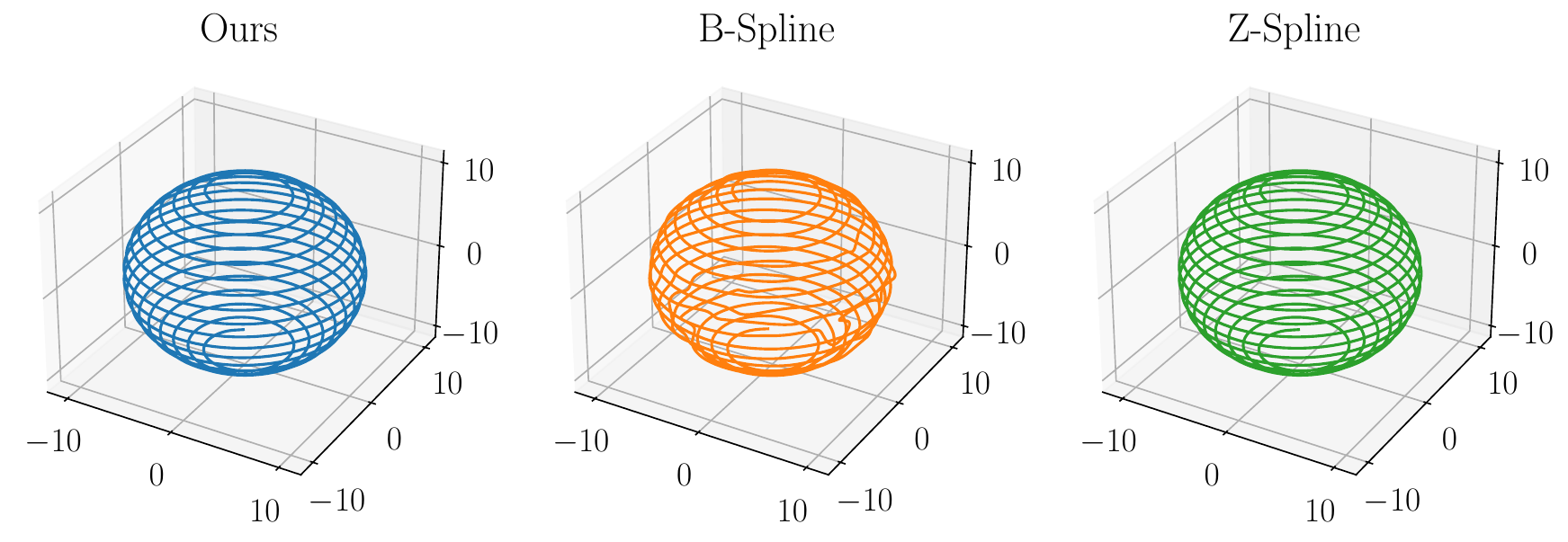}
        \caption{$\eta_{ig}=1, \sigma=10^{-4}$ (both $\mathrm{[m]}$ and $\mathrm{[rad]}$).}
    \end{subfigure}
    \vspace{0.2cm}

    \begin{subfigure}{0.40\textwidth}
        \centering
        \includegraphics[width=\textwidth]{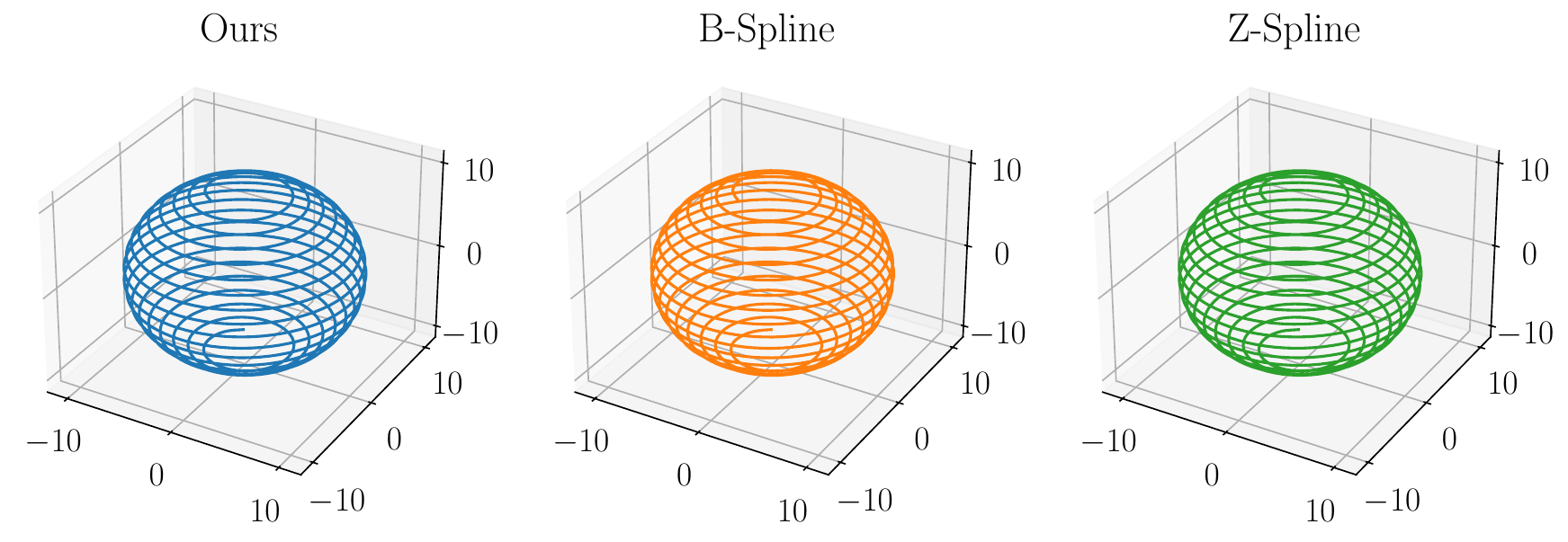}
        \caption{$\eta_{ig}=1, \sigma=10^{-3}$ (both $\mathrm{[m]}$ and $\mathrm{[rad]}$).}
    \end{subfigure}
    \vspace{0.2cm}

    \begin{subfigure}{0.40\textwidth}
        \centering
        \includegraphics[width=\textwidth]{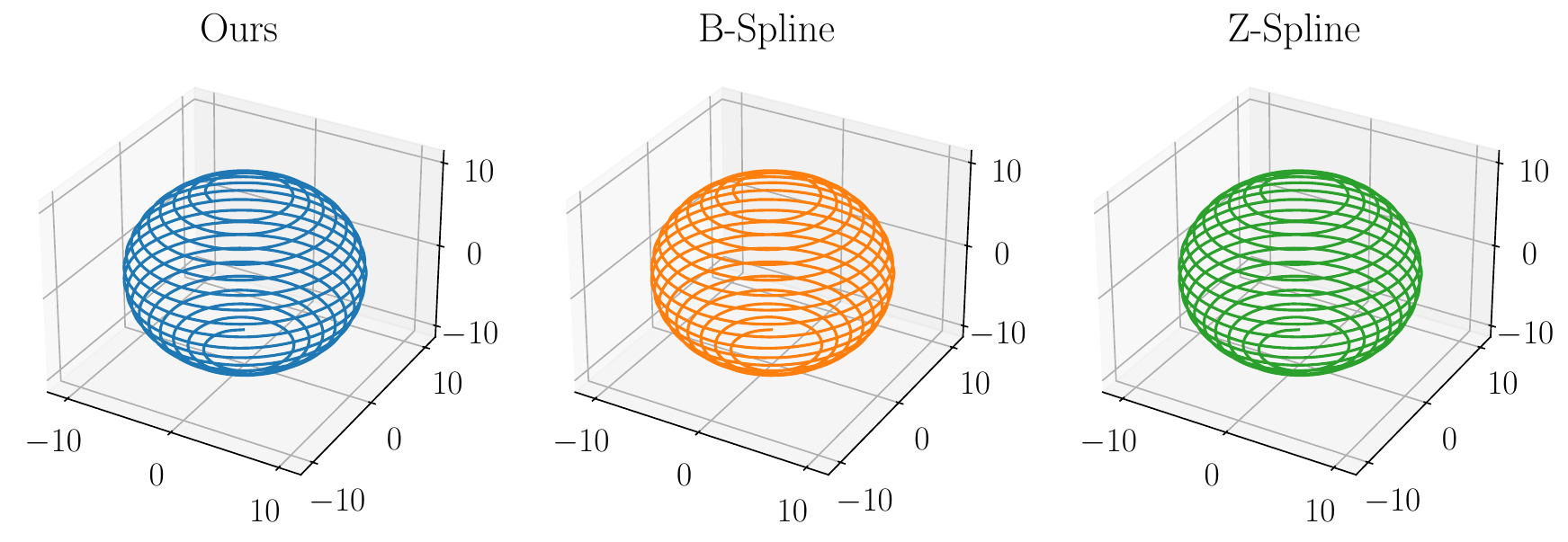}
        \caption{$\eta_{ig}=1, \sigma=10^{-2}$ (both $\mathrm{[m]}$ and $\mathrm{[rad]}$).}
    \end{subfigure}
    \vspace{0.2cm}

    \begin{subfigure}{0.40\textwidth}
        \centering
        \includegraphics[width=\textwidth]{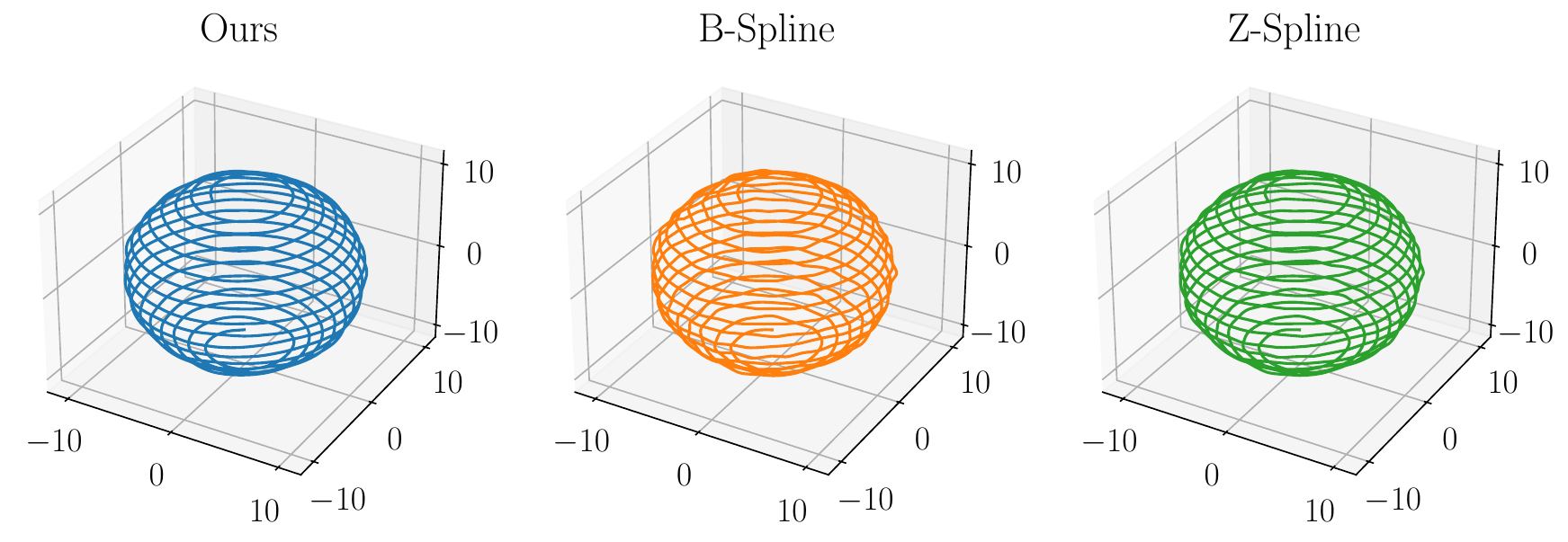}
        \caption{$\eta_{ig}=1, \sigma=10^{-1}$ (both $\mathrm{[m]}$ and $\mathrm{[rad]}$).}
    \end{subfigure}
    \vspace{0.2cm}

    \begin{subfigure}{0.40\textwidth}
        \centering
        \includegraphics[width=\textwidth]{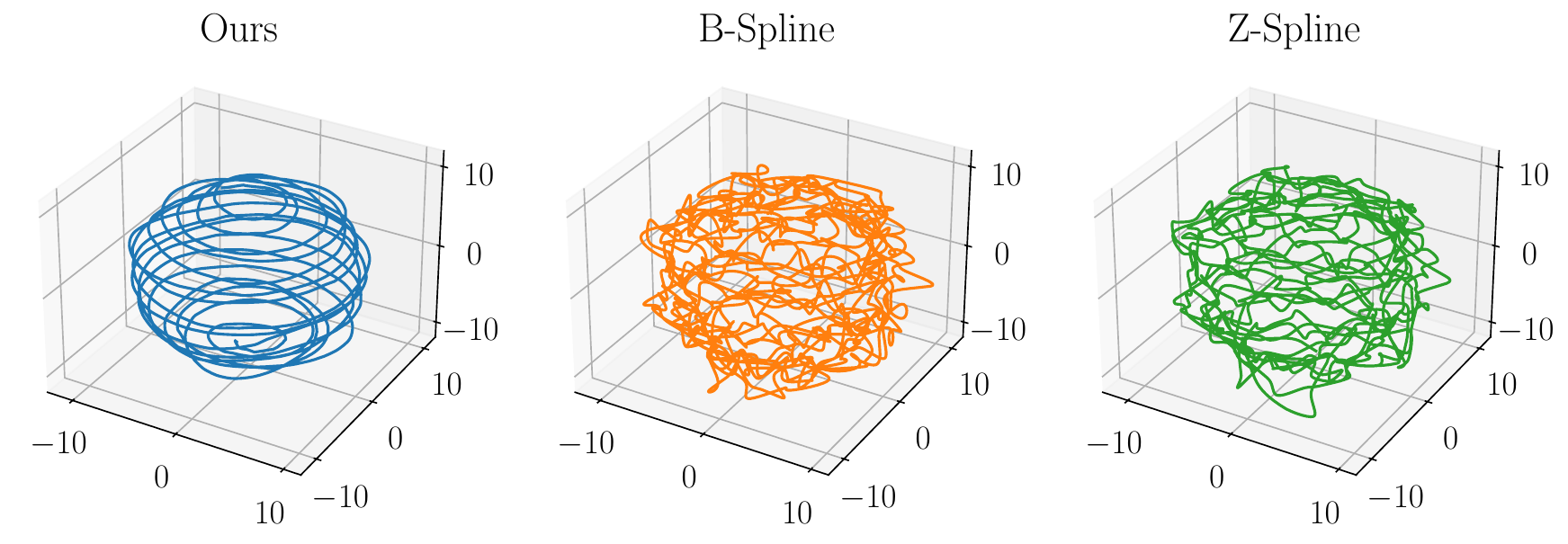}
        \caption{$\eta_{ig}=1, \sigma=1$ (both $\mathrm{[m]}$ and $\mathrm{[rad]}$).}
    \end{subfigure}
    \vspace{0.2cm}

    \begin{subfigure}{0.40\textwidth}
        \centering
        \includegraphics[width=\textwidth]{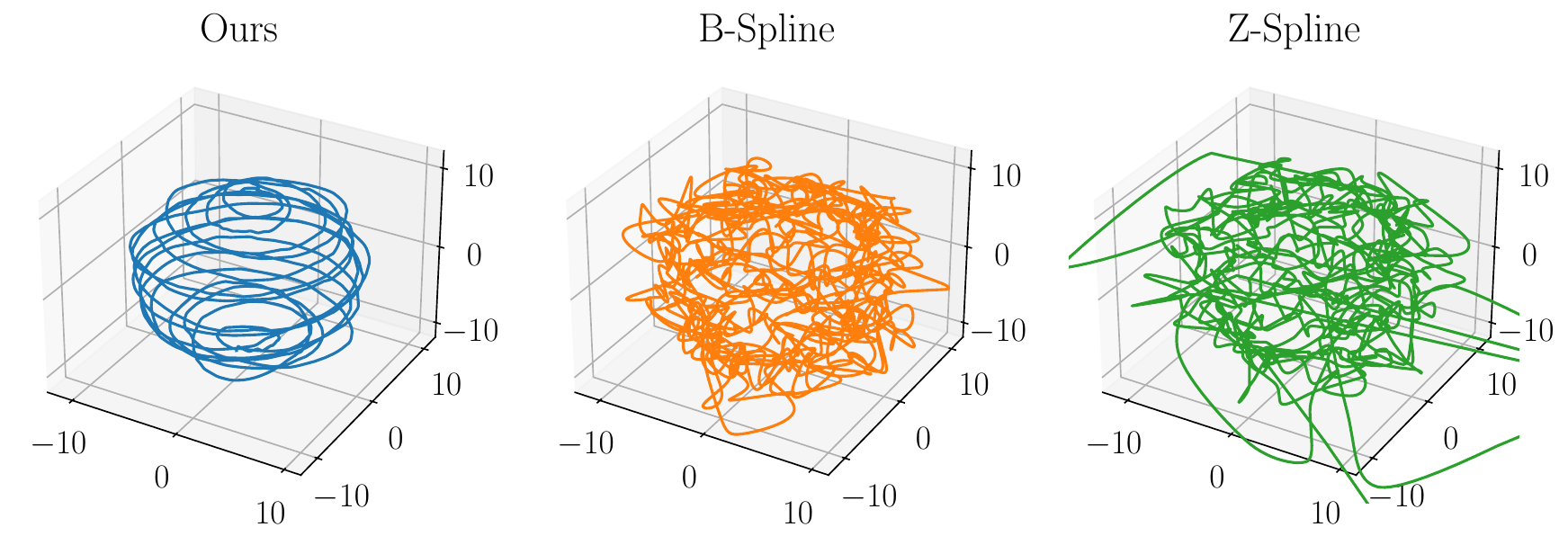}
        \caption{$\eta_{ig}=1, \sigma=1.5$ (both $\mathrm{[m]}$ and $\mathrm{[rad]}$).}
    \end{subfigure}

    \caption{\textbf{PGO experiments.} Comparison between G-solver and Hyperion using the \textit{sphere} trajectory under varying measurement Gaussian noise standard deviations $\sigma$. The initial guess is perturbed with a standard deviation of $\eta_{ig} = 1$ (applied to both positions $\mathrm{[m]}$ and orientations $\mathrm{[rad]}$). All methods query the trajectory at a resolution $100\times$ finer than the control-point spacing.}
    \label{fig:pgo2}
\end{figure}

\begin{figure*}[ht!]
    \centering

    \begin{subfigure}{0.6\textwidth}
        \centering
        \includegraphics[width=\textwidth]{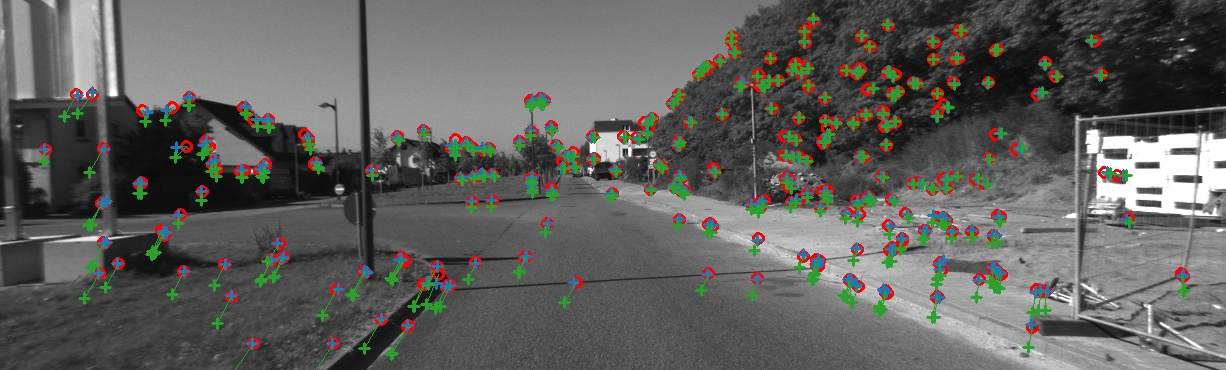}
        \caption{Frame 79.}
    \end{subfigure}
    \vspace{0.3cm}
    
    \begin{subfigure}{0.6\textwidth}
        \centering
        \includegraphics[width=\textwidth]{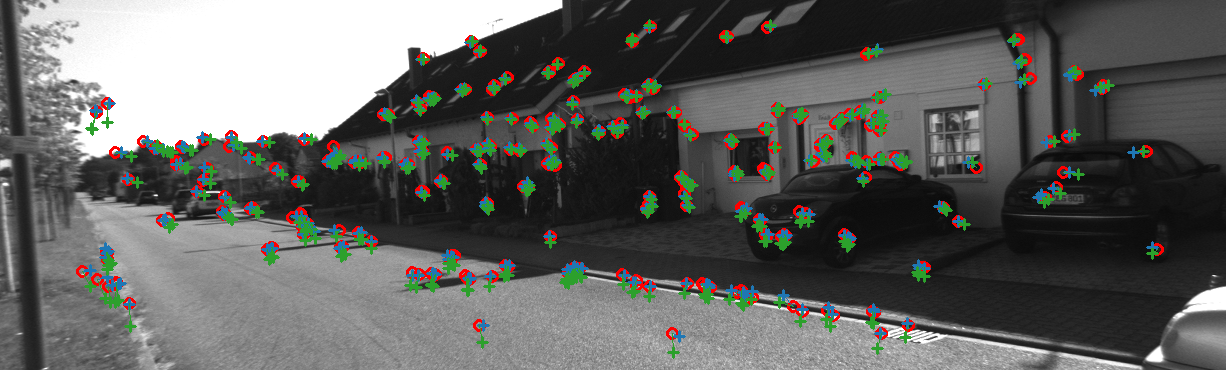}
        \caption{Frame 320.}
    \end{subfigure}
    \vspace{0.3cm}

    \begin{subfigure}{0.6\textwidth}
        \centering
        \includegraphics[width=\textwidth]{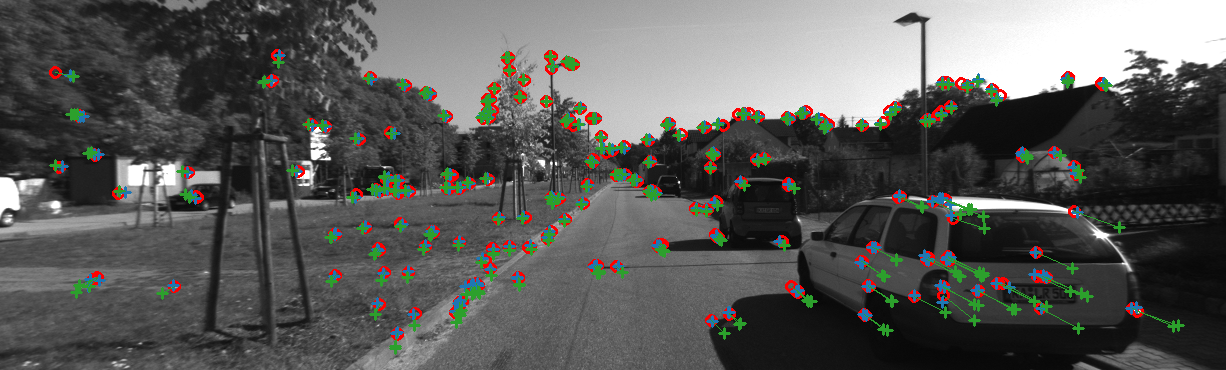}
        \caption{Frame 354.}
    \end{subfigure}
    \vspace{0.3cm}

    \begin{subfigure}{0.6\textwidth}
        \centering
        \includegraphics[width=\textwidth]{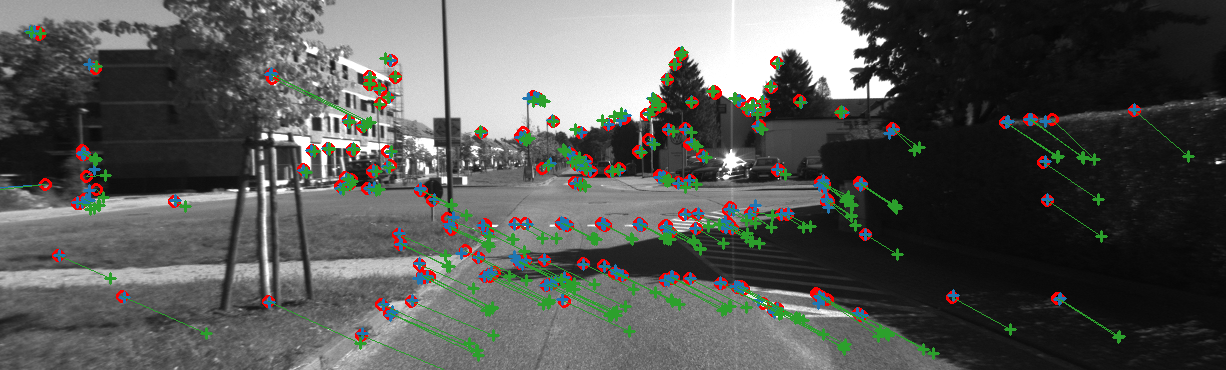}
        \caption{Frame 429.}
    \end{subfigure}
    \vspace{0.3cm}

    \begin{subfigure}{0.6\textwidth}
        \centering
        \includegraphics[width=\textwidth]{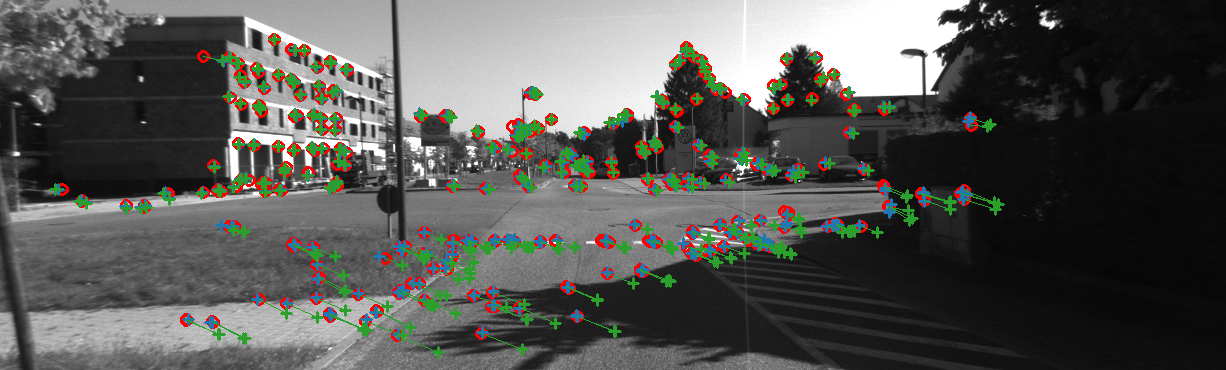}
        \caption{Frame 433.}
    \end{subfigure}
    \vspace{0.3cm}

    \caption{\textbf{Rolling shutter reprojection.} 
    Reprojection of optimized 3D scene landmarks on KITTI~\textit{06} \cite{geiger2013vision} under a $1$\,ms simulated readout time. 
    Hyperion \cite{hyperion} results are shown in \textcolor[rgb]{0.173,0.627,0.173}{green} and G-solver in \textcolor[rgb]{0.121,0.467,0.705}{blue}. 
    Nominal global shutter projections are shown in \textcolor[rgb]{1,0,0}{red}.}
    \label{fig:repr1}
\end{figure*}

\begin{figure*}[ht!]
    \centering

    \begin{subfigure}{0.6\textwidth}
        \centering
        \includegraphics[width=\textwidth]{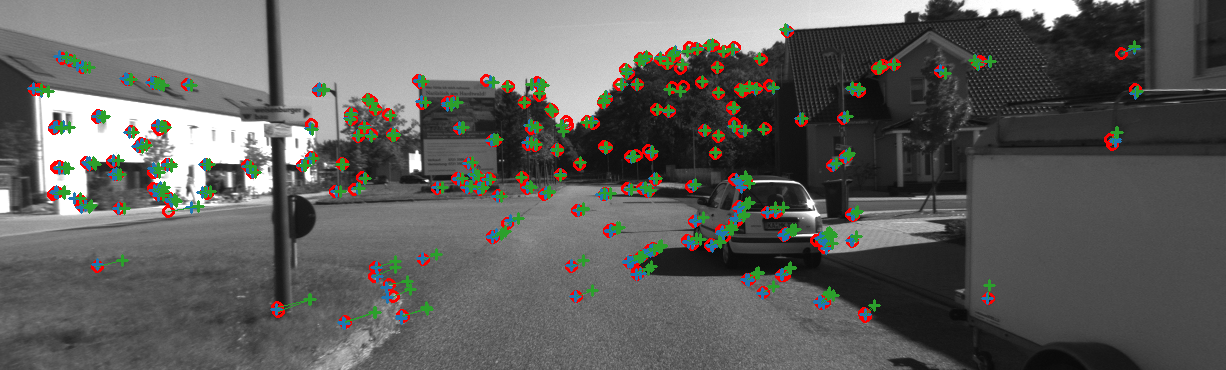}
        \caption{Frame 668.}
    \end{subfigure}
    \vspace{0.3cm}
    
    \begin{subfigure}{0.6\textwidth}
        \centering
        \includegraphics[width=\textwidth]{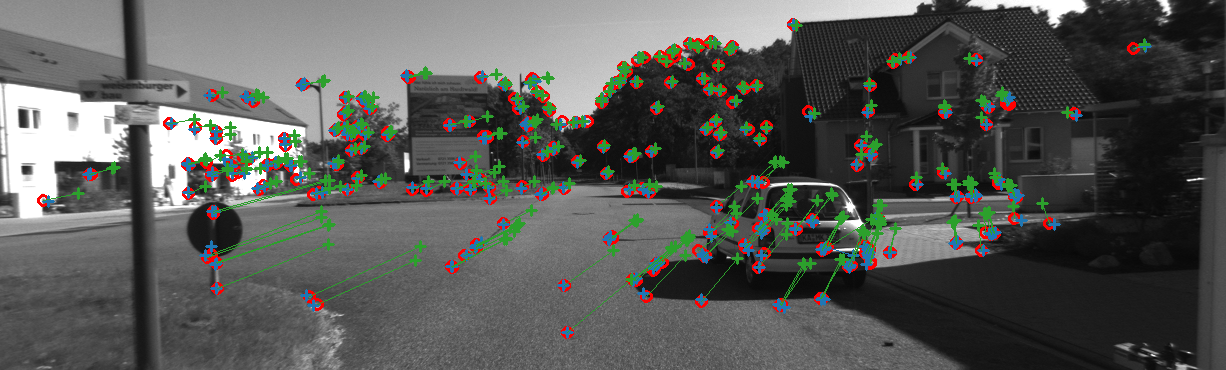}
        \caption{Frame 670.}
    \end{subfigure}
    \vspace{0.3cm}

    \begin{subfigure}{0.6\textwidth}
        \centering
        \includegraphics[width=\textwidth]{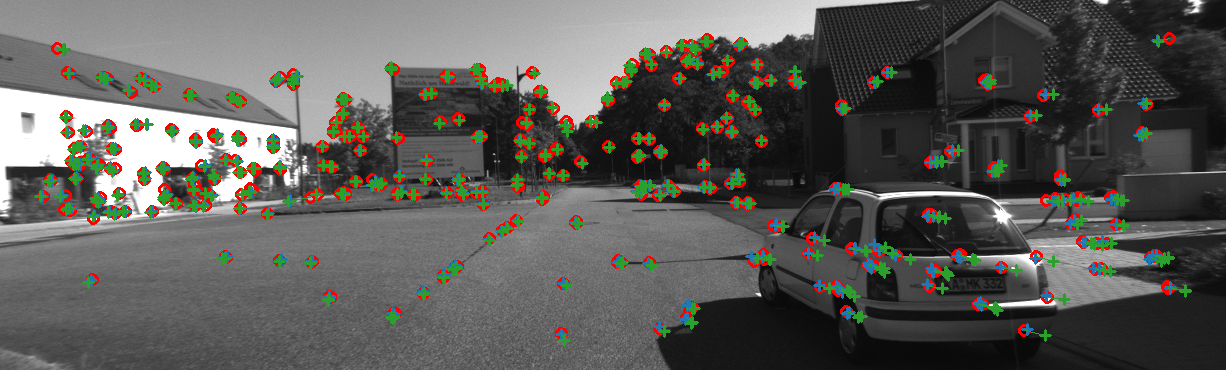}
        \caption{Frame 674.}
    \end{subfigure}
    \vspace{0.3cm}

    \begin{subfigure}{0.6\textwidth}
        \centering
        \includegraphics[width=\textwidth]{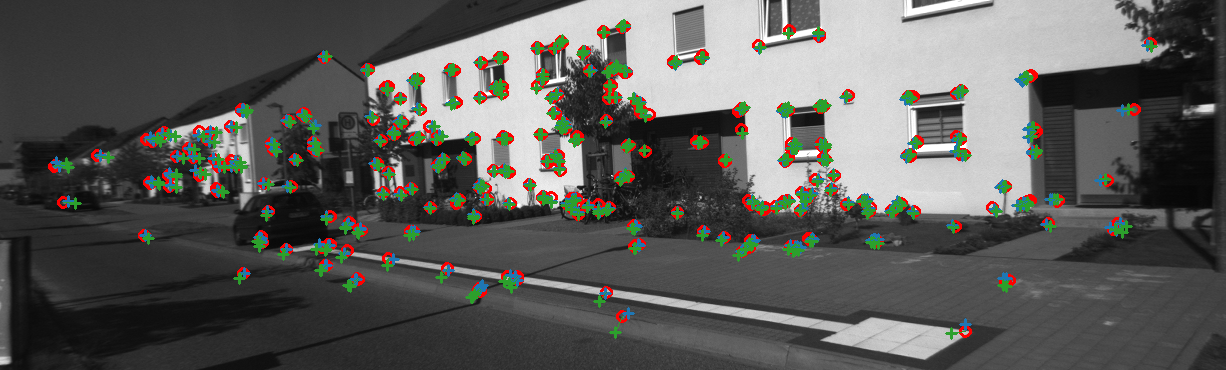}
        \caption{Frame 715.}
    \end{subfigure}
    \vspace{0.3cm}

    \begin{subfigure}{0.6\textwidth}
        \centering
        \includegraphics[width=\textwidth]{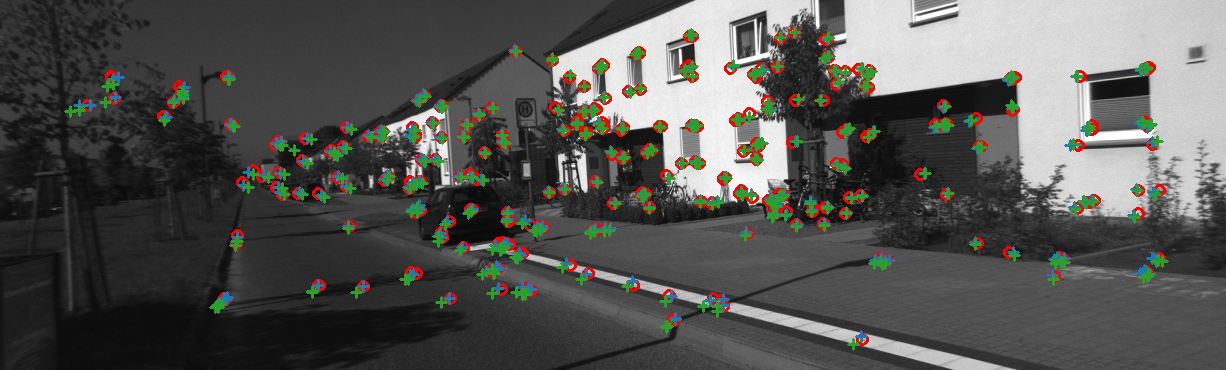}
        \caption{Frame 719.}
    \end{subfigure}
    \vspace{0.3cm}

    \caption{\textbf{Rolling shutter reprojection.} 
    Reprojection of optimized 3D scene landmarks on KITTI~\textit{06} \cite{geiger2013vision} under a $1$\,ms simulated readout time. 
    Hyperion \cite{hyperion} results are shown in \textcolor[rgb]{0.173,0.627,0.173}{green} and G-solver in \textcolor[rgb]{0.121,0.467,0.705}{blue}. 
    Nominal global shutter projections are shown in \textcolor[rgb]{1,0,0}{red}.}
    \label{fig:repr2}
\end{figure*}

Qualitative results are reported for the same synthetic experiments presented in the paper. Prior-based localization using only absolute pose measurements is shown in \figref{fig:prior1} and \figref{fig:prior2}, while \ac{pgo} results are shown in \figref{fig:pgo1} and \figref{fig:pgo2}. After optimization, all the continuous-time methods evaluate the trajectory at a resolution $100\times$ finer than the control-point spacing.
From visual inspection of the trajectories, for noise levels up to a standard deviation of $\sigma=10^{-2}$ (both in $\mathrm{[m]}$ and $\mathrm{[rad]}$), it is difficult to distinguish differences between the estimated solutions. At $\sigma=10^{-1}$, G-solver still provides smooth trajectories, whereas noise clearly starts to degrade Hyperion (both B and Z-spline variants). Finally, for $\sigma=10^{-1}$ and above, G-solver is the only method able to robustly recover the underlying shape despite the strong perturbations.
\\Additional frames from the same rolling shutter experiment presented in the paper are shown in \figref{fig:repr1} and \figref{fig:repr2}. These figures show reprojections of optimized 3D scene landmarks on KITTI~\textit{06} \cite{geiger2013vision}, with a readout time of $1~\mathrm{ms}$. The results demonstrate that our continuous-time modeling produces noticeably sharper and more accurate reprojections compared to Hyperion \cite{hyperion} with Z-Splines.

\clearpage

\end{document}